\documentclass{article}

\usepackage{microtype}
\usepackage{graphicx}
\usepackage{booktabs}
\usepackage{array}
\usepackage{hyperref}

\usepackage[preprint]{icml2026}
\hypersetup{pdfsubject={}}

\usepackage{amsmath}
\usepackage{amssymb}
\usepackage{mathtools}
\usepackage{amsthm}

\theoremstyle{plain}
\newtheorem{theorem}{Theorem}[section]
\newtheorem{proposition}[theorem]{Proposition}
\newtheorem{lemma}[theorem]{Lemma}
\newtheorem{corollary}[theorem]{Corollary}
\theoremstyle{definition}
\newtheorem{definition}[theorem]{Definition}
\newtheorem{assumption}[theorem]{Assumption}
\newtheorem{example}[theorem]{Example}
\theoremstyle{remark}
\newtheorem{remark}[theorem]{Remark}

\newcommand{\E}{\mathbb E}
\newcommand{\Prob}{\mathbb P}
\newcommand{\I}{\mathcal I}

\newcommand{\one}{\mathbf 1}
\newcommand{\Dcl}{\mathcal D}
\newcommand{\Hc}{\mathcal H}

\icmltitlerunning{Clean Sheets and Checkpoints}

\begin{document}

\twocolumn[
  \icmltitle{Clean Sheets and Checkpoints:\\
    Evidence Complexity for Certifying Sequential Agent Risk}

  \icmlsetsymbol{nbl}{\dag}

  \begin{icmlauthorlist}
    \icmlauthor{Rohith Yanambaka Venkata}{indep,nbl}
    \icmlauthor{Don McBride}{indep,nbl}
    \icmlauthor{Tiago Sequeira}{nokia}
    \icmlauthor{Md.\ Al Amin}{indep,nbl}
  \end{icmlauthorlist}

  \icmlaffiliation{indep}{Independent Researcher, United States}
  \icmlaffiliation{nokia}{Cyber Defense Center, Nokia}

  \icmlcorrespondingauthor{Rohith Yanambaka Venkata}{\mbox{yvrohith@gmail.com}}
  \icmlkeywords{AI safety evaluation, certification, sample complexity, renewal theory, dangerous capabilities}

  \vskip 0.3in
]

\printAffiliationsAndNotice{\textsuperscript{\dag}Contributed to this work while employed at Nokia Bell Labs. }

\begin{abstract}
Dangerous-capability evaluations often test an AI agent from prepared states at ordered milestones. A clean sheet
records no successes. What it certifies about a campaign of $B$ attempts depends on what a failed attempt erases.
We model the task as a chain of $k$ milestones divided into checkpoint segments and ask how many zero audit responses
certify, at level $\alpha$, a completion probability of at most $\epsilon$. For segments chosen before the audit,
counting possible completion windows gives a certificate for any specified budget. The bound also covers adaptive
deployment processes that follow the chain's transitions, provided their success probabilities after every complete
history are capped by the corresponding audit success probabilities. The audit does not test this transport
assumption. In rare-risk phase interiors, each segment is either rarely completed within the budget (active) or
nearly always completed (free). The minimum expected number of zero responses is asymptotic to
$\log(1/\alpha)\,(B^a/(a!\,\epsilon))^{1/K_A}$, where the $a$ active segments contain $K_A$ milestones and are those
at least as long as a budget-dependent threshold. The certificate attains this constant. Every asymptotically optimal
evaluator must divide its expected response count equally among active milestones in the limit, with a vanishing share elsewhere. Where a single
segment passes from active to free, each of its milestones receives a smaller limiting share than an active milestone; a
crossover-aware certificate attains the optimal constant. For selected segments with known or unqueried prefix steps,
counting failed cycles can sharpen the certificate.
\end{abstract}

\section{Introduction}\label{sec:intro}

Before an AI agent is deployed, its developer may estimate the probability that it could complete a dangerous
multi-step task, such as setting up its own infrastructure or acquiring resources
\citep{shevlane2023model,phuong2024evaluating}. Complete successes are rare, so the milestone method of
\citet{phuong2024evaluating} divides the task into ordered milestones, starts the agent from a prepared state at the
beginning of each, and runs it a few times there. If every one of these runs fails, the evaluation produces a
\emph{clean sheet}: an all-zero record.

\begin{figure}[t]
  \centering
  \includegraphics[width=\columnwidth]{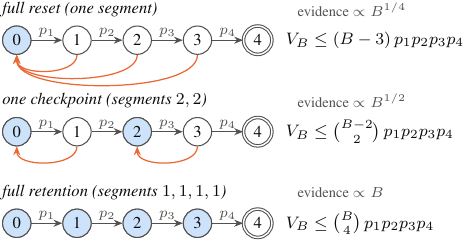}
  \caption{Reset semantics for $k=4$ milestones. A state counts the milestones passed; shaded states start a segment,
  so progress up to them is retained. An attempt from state $i$ succeeds with probability $p_{i+1}$; a failure returns
  the agent to the start of its current segment (arrows; a failure at a segment start leaves the state unchanged).
  Right: the completion-window bound of Theorem~\ref{thm:finite} with every segment selected; above it, the budget factor in the rare-risk
  asymptotic cost (Theorem~\ref{thm:sharp}).}
  \label{fig:semantics}
\end{figure}

A clean sheet is evidence about single attempts from prepared states, whereas a deployment risk statement concerns a
\emph{campaign} of many attempts. What a failed attempt erases decides which stretches of the task must be passed
without a failure, and therefore how quickly the risk grows with the number of attempts and which milestones the audit
must test. If every failure returns the agent to the beginning (full reset), completion needs an uninterrupted run
through all $k$ milestones. If the agent keeps every milestone it has passed (full progress retention), a failure
costs only the attempt. With common milestone probability $p$, the completion probability within $B$ attempts is approximately $Bp^k$
and $(Bp)^k/k!$, respectively. These hold as $B\to\infty$ and $p\to0$, with $Bp^k\to0$ under full reset and $Bp\to0$ under full
retention. Checkpoints that save progress at some milestones lie in between (Figure~\ref{fig:semantics}).

The same record therefore supports different deployment statements. Take four milestones with ten zero responses
each, and a target completion probability of at most $1/100$ at confidence $97.5\%$. Under the assumptions of
Section~\ref{sec:model}, this record meets the target for budgets up to 169 primitive attempts under full reset by an
analytic bound, and up to 185 by a computer-assisted one, but only up to 10 under full retention
(Section~\ref{sec:gdm}).

Suppose each failure erases the progress made in the current checkpoint segment. We ask how many zero audit responses
are needed to certify, with error probability at most $\alpha$, that the agent completes the chain within $B$ charged
attempts with probability at most $\epsilon$. We also ask how those responses should be divided among milestones.
We measure cost by the expected number of responses in the \emph{zero world}, where every milestone success
probability is zero. This is the expected size of the clean sheet collected before certification.

Under the model of Section~\ref{sec:model} the answer has three parts. First, counting completion windows gives a
certificate for any specified budget, including for adaptive deployment processes whose milestone success probabilities
conditional on every complete history are capped by the corresponding audit probabilities (Section~\ref{sec:finite}).
Second, call a segment \emph{active} if its completion within the budget remains rare and \emph{free} if completion
is nearly certain. In a rare-risk \emph{phase interior}, every segment is one or the other. The minimum evidence
is asymptotic to $\log(1/\alpha)(B^a/(a!\,\epsilon))^{1/K_A}$, where $K_A/a$ is the mean length of the $a$ active segments,
so the budget enters with exponent $a/K_A$. The certificate attains this cost (Section~\ref{sec:sharp}). For the rules of
Figure~\ref{fig:semantics}, these exponents are $1/4$, $1/2$ and $1$. Third, every asymptotically optimal evaluator must divide its expected response count equally among
active milestones in the limit, with a vanishing share on free ones. A single segment passing from active to free instead needs a reduced
limiting share per milestone (Section~\ref{sec:crossover}).
Sections~\ref{sec:restricted}--\ref{sec:numerics} add a refinement for known or unqueried steps, numerical
illustrations and a conditional reanalysis of a published clean sheet; proofs are in the appendix.

The ingredients are classical: upper confidence limits for products of binomial parameters from failure-free tests,
Wald's identity and the exponential limit of geometric sums, a log-concavity theorem, and a change-of-measure lower
bound. The contribution is the characterization that connects reset geometry, finite certification of campaign risk,
phase-dependent evidence complexity and the necessary allocation of tests (Section~\ref{sec:related}).

\section{Model}\label{sec:model}

\subsection{Campaigns, milestones and reset semantics}

A task consists of $k\ge1$ mandatory milestones that must be passed in order. A reset rule partitions them into
consecutive \emph{segments} of lengths $\ell_1,\dots,\ell_s\ge1$ with $\sum_j\ell_j=k$; $E_j$ denotes the milestones of
segment $j$, and $L=\max_j\ell_j$. The state $i\in\{0,\dots,k\}$ is the number of milestones passed, and state $k$ is
absorbing. From a state $i<k$ one \emph{charged attempt} is made at milestone $i+1$. A success moves the state to
$i+1$. A failure moves it to $r(i)$, the start of the segment containing milestone $i+1$; $r(i)$ counts the milestones
in all preceding segments. Thus a failure erases progress within the current segment and preserves completed segments. One segment ($s=1$) is
\emph{full reset}, and segments of length one are \emph{full progress retention}. For $p\in[0,1]^k$, the
\emph{independent chain} succeeds at milestone $i$ with probability $p_i$ on every attempt, independently of the past;
$V_B(p)$ is its probability of reaching state $k$ from state $0$ within $B$ attempts, and $\beta_j=\prod_{i\in
E_j}p_i$ is the probability that one uninterrupted pass through segment $j$ succeeds.

The budget $B$ counts primitive charged attempts, one per milestone attempted, not complete trials. Under full reset,
a complete trial runs from the start until its first failure or until completion. A trial that fails at milestone
$i$ therefore consumes $i$ attempts. Statements about trials use a different unit: $n$ independent complete trials
succeed with probability $1-(1-\prod_ip_i)^n$ (Section~\ref{sec:gdm}).

\subsection{Deployment processes}

We allow adaptation, randomization, memory and user guidance subject to the three conditions below. In particular,
the success caps must hold after every complete history (D3).

\begin{assumption}[Domination conditions]\label{ass:D}
A deployment process belongs to $\Dcl(p)$ if
(D1) its only state changes are those of the chain, so that from $i<k$ an attempt either succeeds to $i+1$ or fails
to $r(i)$, no milestone is skipped, no alternative route exists and no progress within a segment survives a failure;
(D2) every attempt consumes one unit of the budget, and the process may stop at any time; and
(D3) for every full history $h$ ending in a state $i<k$, including any memory, randomization, scaffold state or policy
information, $\Prob(\text{next attempt succeeds}\mid h)\le p_{i+1}$.
\end{assumption}

\begin{lemma}[Domination]\label{lem:domination}
$V_B(p)$ is nondecreasing in $B$ and in each coordinate of $p$. Every process in $\Dcl(p)$ completes within $B$
attempts with probability at most $V_B(p)$, and the independent chain belongs to $\Dcl(p)$.
\end{lemma}

The proof (Appendix~\ref{app:model}) is a backward induction on the remaining budget, uniform over histories; it
specializes robust dynamic programming \citep{iyengar2005robust,nilim2005robust} to the chain (compare Proposition~2.1
of \citet{liu2026certified}, using a lower bound on per-turn safety that holds in every safe dialogue state). Condition (D3) cannot be weakened to a cap on averages: if every attempt succeeds exactly when a campaign-level latent variable
$Z\sim\mathrm{Bernoulli}(r)$ equals one, an attempt at either milestone from a fresh campaign succeeds with
probability $r$, yet a two-milestone campaign with full retention completes within two attempts with probability
$r>V_2(r,r)=r^2$ (Example~\ref{ex:average}).

\subsection{Audit observations and validity}

\begin{assumption}[Audit oracle]\label{ass:O}
An evaluator may repeatedly choose a milestone $i$ and receive a response that, conditionally on everything observed
so far and on the evaluator's private randomization, is Bernoulli$(p_i)$. Choices and stopping may be adaptive. The
cost of constructing or reaching milestone states is not counted.
\end{assumption}

Let $N_i$ count the responses at milestone $i$, $N=\sum_iN_i$, and $c_\alpha=\log(1/\alpha)$ for
$\epsilon,\alpha\in(0,1)$. An evaluator is \emph{valid} at $(B,\epsilon,\alpha)$ if
\[
  \sup_{p:\,V_B(p)>\epsilon}\Prob_p\bigl(\text{it ever certifies }V_B(p)\le\epsilon\bigr)\le\alpha
\]
and, when $p=0$, it certifies after finitely many responses almost surely. Nothing is required at other safe
parameters, where an evaluator may return ``unresolved''. Its \emph{zero-world cost} is $\E_0N$, the expected number
of responses when every $p_i=0$.

The zero-world restriction applies to evidence-complexity optimality, not to the campaign-risk bounds:
Theorem~\ref{thm:finite}(i) holds for every $p\in[0,1]^k$ and every process in $\Dcl(p)$. At $p=0$, every audit
response is zero, so $\E_0N$ measures the expected size of the clean sheet collected before certification. Minimizing
this cost asks how large a clean sheet must be.

Validity concerns the audit parameter $p$. A statement about deployment needs, in addition, that the deployment process
lies in $\Dcl(p)$ for the same $p$, that is, that success probabilities measured from prepared milestone states cap
those of the states reached in deployment at every history. We call this the \emph{transport assumption}. The audit
does not test it; it links the results about $\Dcl(p)$ to a deployment.

\subsection{The certification geometry}

For a fixed-count design that certifies on an all-zero record, validity requires that this record have probability at most
$\alpha$ under every unsafe parameter. The hardest unsafe parameter is therefore the one under which that record
is most likely.

An all-zero record with $n_i$ responses at milestone $i$ has probability $\exp(-\sum_in_iu_i)$ under $p$, where
$u_i=-\log(1-p_i)$, so in these coordinates the log-likelihood of a clean sheet is linear. For $u\in[0,\infty)^k$ let
$F_B(u)=V_B(\one-e^{-u})$ and
\begin{gather*}
  U=\{u:F_B(u)>\epsilon\},\qquad S=\{u:F_B(u)\ge\epsilon\},\\
  I_w=\inf_{u\in S}w^\top u\quad(w\in\Delta_k),\qquad \I^*=\max_{w\in\Delta_k}I_w ,
\end{gather*}
with $\Delta_k$ the probability simplex.

Each $u_i=\mathrm{KL}(\mathrm{Ber}(0)\,\|\,\mathrm{Ber}(p_i))$ is the
log-likelihood ratio that one zero response at milestone $i$ contributes against $p$, so a design that places a
fraction $w_i$ of $N$ responses at milestone $i$ and certifies when all are zero errs at $u\in U$ with probability
$\exp(-Nw^\top u)$. Thus $I_w$ is the information per response against the hardest unsafe parameter, and the design
needs $c_\alpha/I_w$ responses (with real counts). By Lemma~\ref{lem:zeropath} and Proposition~\ref{prop:sandwich},
$c_\alpha/\I^*$ is the minimum zero-world cost of any valid evaluator, adaptive or not, up to an additive $k$, and the
maximizing allocation says how to spread a clean sheet over the milestones.

\begin{lemma}[Closure and diagonal]\label{lem:closure}
Let $B\ge k$. Then $S$ is the closure of $U$, there is a unique $t>0$ with $F_B(t\one)=\epsilon$, $\I^*$ is attained,
and $\I^*\le t$.
\end{lemma}

We call $t$ the \emph{diagonal threshold}; $p_*=1-e^{-t}$ is the common milestone probability at which the campaign
risk equals $\epsilon$.

Neither $S$ nor its image in logit coordinates need be convex: Proposition~\ref{prop:nonconvex} gives interior
counterexamples under full reset with five milestones, several of them nearly certain, at $\epsilon=1/100$ in logit and
$\epsilon\approx0.02$ in $u$ coordinates. Our prospective certificates bound $V_B$ by functions that are log-concave in
$u$, and no result here assumes that $S$ is convex.

\section{Finite certification under different reset rules}\label{sec:finite}

Completing the chain requires one uninterrupted run of successes through every segment, and counting the possible end
positions of these runs bounds the completion probability. For a nonempty set $A$ of segments let $a=|A|$,
$E_A=\bigcup_{j\in A}E_j$, $K_A=|E_A|$, and for $B\ge k$
\begin{gather*}
  C_A(B)=\binom{B-k+a}{a},\\
  r_A=\bigl(\epsilon/C_A(B)\bigr)^{1/K_A},\qquad \tau_A=-\log(1-r_A).
\end{gather*}

\begin{theorem}[Finite selected-segment certificate]\label{thm:finite}
Let $B\ge k$ and let $A$ be a nonempty set of segments.
\begin{enumerate}
\item[(i)] For every $p\in[0,1]^k$ and every process in $\Dcl(p)$,
\[
  \Prob(\text{completion within }B)\le V_B(p)\le C_A(B)\prod_{i\in E_A}p_i .
\]
\item[(ii)] $\tau_A\le I_{w_A}\le\I^*\le t$, where $w_A$ is uniform on $E_A$.
\item[(iii)] Suppose that $A$ is fixed before any response is observed; it may depend on $k$, the segment lengths,
$B$, $\epsilon$, $\alpha$ and randomization independent of the responses. Query each milestone of $E_A$ exactly
$n_A=\lceil c_\alpha/(K_A\tau_A)\rceil$ times and certify $V_B(p)\le\epsilon$ if and only if all responses are zero.
This rule is valid, certifies with probability one at $p=0$, and uses $N=K_An_A<c_\alpha/\tau_A+K_A$ responses.
Milestones outside $E_A$ are never queried, and their parameters are unrestricted.
\end{enumerate}
\end{theorem}

For (i), let $b_m$ be the attempt at which the $m$-th selected segment is completed and $\bar k_m$ the number of
milestones up to its end. Then $x_m=b_m-\bar k_m$ counts the attempts spent beyond the minimum. Mandatory progress
gives $0\le x_1\le\dots\le x_a\le B-k$, and there are $\binom{B-k+a}{a}$ such sequences. Each sequence fixes disjoint
windows for the final uninterrupted runs through the selected segments. By (D3), all these runs succeed with
probability at most $\prod_{E_A}p_i$. A union bound over sequences gives (i). The count uses (D1)--(D3) but not
independence. Parts (ii) and (iii) follow from the concavity of $u\mapsto\log(1-e^{-u})$ (Appendix~\ref{app:finite}).

The certificate holds at every $(B,\epsilon,\alpha)$, for adaptive deployment processes and arbitrary unqueried
milestones. Under full reset it gives $V_B(p)\le(B-k+1)\prod_ip_i$ and about $c_\alpha((B-k+1)/\epsilon)^{1/k}$ zero
responses; under full retention, $V_B(p)\le\binom{B}{k}\prod_ip_i$ and a cost of order $c_\alpha B(k!\,\epsilon)^{-1/k}$.
Choosing the $A$ that minimizes $K_An_A$ uses only the model and $(B,\epsilon,\alpha)$, so the choice is prospective
and the rule remains valid.

Given (i), part (iii) is an upper confidence limit for $\prod_{E_A}p_i$ from an all-zero outcome with equal counts.
One-sided confidence limits for a product of two binomial parameters were studied by \citet{buehler1957confidence};
failure-free component tests propagated to system bounds,
with unequal, zero or equal allocation, are developed by \citet{bishop2020conservative,bishop2022optimising}. What is specific to the chain is the completion-window
coefficient $C_A(B)$, its domination of every process in $\Dcl(p)$, and the prospective choice of segments.

\begin{remark}[Conditions]\label{rem:finite}
(a) If $A$, the counts or the acceptance rule are chosen after responses are seen, or if the rule may also certify
after nonzero responses, the combined rule needs its own calibration. (b) The window count is exact, but the union
bound can be loose when the selected parameters or the products $B\beta_j$ are not small (Section~\ref{sec:numerics}).
\end{remark}

For counts fixed in advance but not designed for a particular $(B,\epsilon)$, the following rule, used in
Section~\ref{sec:gdm}, certifies all pairs at once (Appendix~\ref{app:finite}).

\begin{proposition}[Fixed-count plausibility rule]\label{prop:plausibility}
Fix counts $n\in\{0,1,\dots\}^k$ before observing data and query milestone $i$ exactly $n_i$ times. On an all-zero
record let $\Pi=\{u\in[0,\infty]^k:\sum_in_iu_i\le c_\alpha\}$, with $0\cdot\infty=0$ and $p_i=1$ when $u_i=\infty$, and certify every
pair $(B,\epsilon)$ with $\sup_{u\in\Pi}F_B(u)\le\epsilon$; after any nonzero response certify nothing. Then for every
$p$, $\Prob_p(\text{some certified pair has }V_B(p)>\epsilon)\le\alpha$.
\end{proposition}

\section{Sharp zero-world evidence complexity and phase-dependent allocation}\label{sec:sharp}

\subsection{Finite information bounds}

\begin{lemma}[Zero-path lower bound]\label{lem:zeropath}
Every valid evaluator satisfies $\E_0N\ge c_\alpha/\I^*$. If $T=\E_0N<\infty$ and $w_i=\E_0N_i/T$, then
$T\,I_w\ge c_\alpha$.
\end{lemma}

Lemma~\ref{lem:zeropath} is the zero-null form of the change-of-measure lower bound of
\citet[Lemma~1]{kaufmann2016complexity}; see also \citet{garivier2016optimal}. That lemma assumes mutually absolutely
continuous arm distributions, which fails here because $\mathrm{Ber}(0)$ gives no mass to a success, so we prove the
bound directly (Appendix~\ref{app:sharp}): conditionally on the evaluator's randomization, an unsafe parameter
reproduces the all-zero path with probability $\exp(-\sum_iu_iN_i)$, and Jensen's inequality gives the claim.

\begin{proposition}[Finite sandwich]\label{prop:sandwich}
Let $B\ge k$ and let $w^*$ attain $\I^*$. The fixed-count all-zero design with $n_i=\lceil w^*_ic_\alpha/\I^*\rceil$
is valid, so that $c_\alpha/\I^*\le\inf\E_0N<c_\alpha/\I^*+k$, the infimum being over valid evaluators. The
information also satisfies $t/k\le I_{\one/k}\le\I^*\le t$.
\end{proposition}

The information optimum $\I^*$ determines the minimum zero-world cost to within fewer than $k$ responses.
The diagonal threshold $t$ gives simpler bounds that differ by a factor of at most $k$, apart from rounding.
Neither gives an explicit leading constant or allocation.

\subsection{Cycle lengths}

A \emph{cycle} starts at the beginning of a segment and ends at the first failure or at segment completion. For segment $j$ let $\pi_{j,h}$ be the product of its first $h$ milestone probabilities
($\pi_{j,0}=1$) and $\kappa_j=\sum_{h=0}^{\ell_j-1}\pi_{j,h}\in[1,\ell_j]$. A cycle has mean length $\kappa_j$ and
completes the segment with probability $\beta_j$, so by Wald's identity the number $T_j$ of attempts spent in segment
$j$ has $\E T_j=\kappa_j/\beta_j$ when $\beta_j>0$. In the independent chain, different segments use disjoint sets of independent attempts. Their attempt counts $T_j$
are therefore independent, and $V_B(p)=\Prob(\sum_jT_j\le B)$.

\begin{lemma}[Cycle-length renewal bounds]\label{lem:renewal}
Let $A$ be a nonempty set of segments.
(a) For every $p\in[0,1]^k$ and $B\ge k$,
$V_B(p)\le\frac{(B-k+K_A)^a}{a!}\prod_{j\in A}\frac{\beta_j}{\kappa_j}$; a matching lower bound is in
Lemma~\ref{lem:renewal-full}.
(b) Along any sequence with fixed segment lengths and fixed $A$ such that $B\to\infty$, $\beta_j>0$ for $j\in A$,
$B\max_{j\in A}\beta_j\to0$ and $\sum_{j\notin A}\kappa_j/(B\beta_j)\to0$,
\[
  V_B(p)\sim\frac{B^a}{a!}\prod_{j\in A}\frac{\beta_j}{\kappa_j}.
\]
No condition is placed on individual milestone probabilities, any of which may stay fixed, including at one.
\end{lemma}

The identity $\E T_j=\kappa_j/\beta_j$ and the rate $\beta_j/\kappa_j$ are classical
\citep{wald1944cumulative}, \citep[Ch.~XIII]{feller1968introduction}; the exponential limit of
$\beta_jT_j/\kappa_j$ is R\'enyi's theorem on geometric sums \citep{renyi1956characterization}, with rarity and
exponentiality treated by \citet{keilson1979markov} and error bounds by \citet{brown1990error} and
\citet{kalashnikov1997geometric}. Lemma~\ref{lem:renewal} uses only first moments, via Wald's identity
with bounded overshoot \citep[cf.][]{lorden1970excess}.

\subsection{Phase interiors}

At a finite budget the unsafe set has no simple shape, and Proposition~\ref{prop:sandwich} identifies $\I^*$ only
up to a factor $k$. The shape simplifies when the budget separates the segments into two kinds: \emph{active} segments,
whose completion within the budget is still rare, and \emph{free} ones, whose completion is nearly certain. Then
Lemma~\ref{lem:renewal}(b) gives $V_B\sim(B^a/a!)\prod_{j\in A}\beta_j/\kappa_j$ over the active set $A$, the boundary
of the unsafe set is asymptotically a level set of $\prod_{i\in E_A}p_i$, and the max--min problem defining $\I^*$ has
an explicit asymptotic solution. A segment that is neither (a crossover) contributes a nondegenerate factor
(Section~\ref{sec:crossover}). On the diagonal, where every milestone probability equals $p$, segment $j$ is active
when $Bp^{\ell_j}$ is small and free when it is large.

\begin{definition}[Diagonal phase-interior sequence]\label{def:phase}
Fix the segment lengths and a nonempty set $A$ of segments. A sequence $(B_n,p_n)$ with integers $B_n\ge k$,
$B_n\to\infty$, and $p_n\in(0,1)$, $p_n\to0$, is a \emph{diagonal phase-interior sequence for $A$} if
$\max_{j\in A}B_np_n^{\ell_j}\to0$ and $\sum_{j\notin A}(B_np_n^{\ell_j})^{-1}\to0$. Put $\epsilon_n=V_{B_n}(p_n\one)$
and $t_n=-\log(1-p_n)$, and index $t$, $\tau_A$, $I_w$ and $\I^*$ by $(B_n,\epsilon_n)$; by Lemma~\ref{lem:closure},
$p_n$ is then the diagonal threshold and $t=t_n$.
\end{definition}

\begin{lemma}[Active sets are length thresholds]\label{lem:active}
If a diagonal phase-interior sequence for $A$ exists, then $A=\{j:\ell_j\ge\ell^*\}$ with
$\ell^*=\min_{j\in A}\ell_j$. Conversely, let $\ell^*$ be a segment length and $\ell_-$ the largest smaller length
($\ell_-=0$ if there is none); for every $\gamma\in(\ell_-,\ell^*)$ and every $p_n\downarrow0$, the budgets
$B_n=\lceil p_n^{-\gamma}\rceil$ give, for large $n$, a diagonal phase-interior sequence for
$\{j:\ell_j\ge\ell^*\}$.
\end{lemma}

A segment of length $\ell$ needs about $p^{-\ell}$ attempts: it is free once $B\gg p^{-\ell}$ and active while
$B\ll p^{-\ell}$, so longer segments stay active longest and there is one phase interior per distinct length.

\begin{theorem}[Sharp zero-world evidence complexity]\label{thm:sharp}
Let $(B_n,p_n)$ be a diagonal phase-interior sequence for $A$. Then $\epsilon_n\to0$,
\[
  \epsilon_n\sim\frac{B_n^ap_n^{K_A}}{a!},\qquad
  \tau_{A,n}\sim\I^*_n\sim t_n\sim\Bigl(\frac{a!\,\epsilon_n}{B_n^a}\Bigr)^{1/K_A}.
\]
If $\alpha_n\in(0,1)$ satisfies $c_{\alpha_n}/t_n\to\infty$ (in particular if $\alpha$ is fixed), then
\[
  \inf_{\text{valid at }(B_n,\epsilon_n,\alpha_n)}\E_0N\;\sim\;c_{\alpha_n}\Bigl(\frac{B_n^a}{a!\,\epsilon_n}\Bigr)^{1/K_A},
\]
and the certificate of Theorem~\ref{thm:finite} with the same $A$ attains this leading constant.
\end{theorem}

The theorem identifies the constant, not only the rate; the exponent $a/K_A$ falls from $s/k$, when every segment is
active, to $1/L$, when only the longest remain. The condition $c_{\alpha_n}/t_n\to\infty$ makes the rounding term $K_A$
negligible; otherwise no constant is claimed.

\begin{theorem}[Necessary limiting allocation]\label{thm:alloc}
Under the hypotheses of Theorem~\ref{thm:sharp}:
\begin{enumerate}
\item[(a)] every sequence $w_n\in\Delta_k$ with $I_{w_n,n}/t_n\to1$ satisfies $w_{n,i}\to1/K_A$ for $i\in E_A$ and
$w_{n,i}\to0$ for $i\notin E_A$;
\item[(b)] for every sequence of valid evaluators with $\E_0N\sim c_{\alpha_n}/t_n$, the allocation weights
$w_{n,i}=\E_0N_i/\E_0N$ satisfy the conclusion of (a).
\end{enumerate}
\end{theorem}

Each weight is the expected count at a milestone divided by the expected total count, both in the zero world.
These weights tend to zero on free milestones and to $1/K_A$ on active ones. Valid evaluators with any other limiting weights are not asymptotically optimal. The statement does not concern realized fractions or finite $n$; at a finite budget the optimal weights need not be uniform (Example~\ref{ex:nonuniform}). Equal and zero
allocations already arise in optimal component-test plans \citep{bishop2022optimising}; the theorem shows that in the
chain they are necessary in the limit, with the reset rule and the budget deciding which milestones receive weight.

Under full reset with $B_np_n^k\to0$ the constant is $c_\alpha(B_n/\epsilon_n)^{1/k}$, and under full retention with
$B_np_n\to0$ it is $c_\alpha B_n(k!\,\epsilon_n)^{-1/k}$, with uniform limiting allocations in both cases
(Corollary~\ref{cor:corners}). Under full retention a finite statement also holds: $F_B$ is Schur-concave in $u$, so $I_{\one/k}=\I^*=t$ at every
budget (Proposition~\ref{prop:schur}). Majorization properties of convolutions of geometric random variables are
studied by \citet{boland1994schur} and \citet{mao2010ordering}; we give a self-contained proof and claim no novelty for
this statement.

\subsection{A single crossover}\label{sec:crossover}

When exactly one segment crosses from active to free, the optimal evidence allocation differs from that in either
neighboring phase interior. Completing this segment takes a number of attempts comparable to the budget. Its
milestones need fewer responses than active ones but not none, and no certificate of Theorem~\ref{thm:finite} is sharp.

When a segment of length $m$ crosses over, it consumes a random fraction of the budget. Its attempt count divided
by $B_n$ tends in law to $E/\lambda$, with $E\sim\mathrm{Exp}(1)$ and $\lambda=\lim B_np_n^m$. The active segments
must fit in the remaining budget, so the risk acquires the factor
$h_a(\lambda)=\E(1-E/\lambda)_+^a$. This factor increases from $0$ to $1$; for example,
$h_1(\lambda)=1-(1-e^{-\lambda})/\lambda$.

The elasticity of $h_a$, $e_a(\lambda)=\lambda h_a'(\lambda)/h_a(\lambda)$, determines the allocation ratio. In the limiting
risk expression, each active milestone has elasticity $1$ and each crossover milestone has elasticity $e_a(\lambda)$.
By Pr\'ekopa's theorem, $e_a$ is nonincreasing, from $1$ to $0$ (Lemma~\ref{lem:profile}).

\begin{theorem}[Single crossover]\label{thm:crossover}
Let $M$ be the only segment of length $m$, with milestones $E_M$, let $A=\{j:\ell_j>m\}\neq\emptyset$, and let
$(B_n,p_n)$ be as in Definition~\ref{def:phase} but with $B_np_n^m\to\lambda\in(0,\infty)$ in place of its two
conditions.
\begin{enumerate}
\item[(i)] The risk and information satisfy
\[
  \epsilon_n\sim\frac{B_n^ap_n^{K_A}}{a!}\,h_a(\lambda),\qquad \I^*_n\sim t_n.
\]
\item[(ii)] If $c_{\alpha_n}/t_n\to\infty$, the minimum zero-world cost is asymptotic to
\[
  c_{\alpha_n}\Bigl(\frac{B_n^a\,h_a(\lambda)}{a!\,\epsilon_n}\Bigr)^{1/K_A},
\]
attained by the certificate of Proposition~\ref{prop:crosscert}.
\item[(iii)] Under the condition in (ii), every sequence of valid evaluators with
$\E_0N\sim c_{\alpha_n}/t_n$ has allocation weights $\E_0N_i/\E_0N$ tending to $\omega(\lambda)$. These weights are proportional to
$1$ on each milestone of $E_A$ and to $e_a(\lambda)$ on each milestone of $E_M$, and are $0$ elsewhere.
\end{enumerate}
\end{theorem}

\begin{proposition}[Crossover-aware certificate]\label{prop:crosscert}
Let $B\ge k$, fix $A$ and a segment $M\notin A$ of length $m$ before any response, and put $B_\times=B-k+a+1$.
\begin{enumerate}
\item[(i)] For every $p$ and every process in $\Dcl(p)$, the completion probability is at most
\[
  V_B(p)\le\frac{B_\times^a}{a!}\,h_a\Bigl(B_\times\prod_{E_M}u_i\Bigr)\prod_{E_A}p_i.
\]
\item[(ii)] Let $\tau_\times$ solve
\[
  (1-e^{-\tau})^{K_A}\frac{B_\times^a}{a!}\,h_a(B_\times\tau^m)=\epsilon.
\]
Let $w$ be proportional to $1/(e^{\tau_\times}-1)$ on $E_A$ and to
$e_a(B_\times\tau_\times^m)/\tau_\times$ on $E_M$, with zero weights elsewhere. Query each $i\in E_A\cup E_M$ exactly
$\lceil c_\alpha w_i/\tau_\times\rceil$ times and certify if and only if all responses are zero. This rule is valid,
certifies with probability one at $p=0$, and uses fewer than $c_\alpha/\tau_\times+K_A+m$ responses. Moreover,
$\tau_\times\le I_w\le\I^*\le t$.
\end{enumerate}
\end{proposition}

The bound (i) is log-concave in $u$, and $w$ is the normalized gradient of its logarithm at the diagonal point. As
$\lambda$ grows, $\omega(\lambda)$ moves from the uniform allocation on $E_A\cup E_M$ to that on $E_A$; for segments
$(3,1)$ at $\lambda=2$, $e_1(2)\approx0.52$, so the length-one segment's milestone receives about half the limiting
expected share of each active milestone. Even the best certificate of Theorem~\ref{thm:finite} costs more at a
crossover; for $(3,1)$, its largest limiting cost ratio to the optimum is about $1.18$
(Corollary~\ref{cor:crosswindow}). The crossover-aware certificate is not uniformly better
at finite budgets, so one uses the smallest prospective certificate (Figure~\ref{fig:complexity}). The diagonal risk
limit is not new (Appendix~\ref{app:crossover}).

\begin{remark}[Scope]\label{rem:scope}
Theorems~\ref{thm:sharp}--\ref{thm:crossover} are asymptotic. They do not identify the optimum at finite
$(B,\epsilon)$ or cover crossovers of several equal-length segments, crossovers without a positive finite limit of
$B_np_n^m$, fixed $\epsilon$ (Example~\ref{ex:fixedeps}), nonzero safe parameters, or realized adaptive allocations.
These limitations do not affect the validity of the finite certificates. Cycle lengths do not affect the stated
constants: the proofs use only points where every $\kappa_j\to1$.
\end{remark}

\section{Cycle-aware certification}\label{sec:restricted}

Evaluation tasks often contain steps whose success probability is known, such as scripted setup actions that always
succeed; under full reset, a later failure also wastes the attempts spent on them. For a segment of length two with probabilities $(1,q)$, every cycle has length two
and $V_B=1-(1-q)^{\lfloor B/2\rfloor}$, so $V_B\sim Bq/2$ as $B\to\infty$ with $Bq\to0$: half the window coefficient $B-1$
(Example~\ref{ex:cyclecost}). Counting cycles instead of windows recovers this factor.

Suppose that some milestones are not queried, so that their success probabilities are bounded only by one. Before
any response, choose a nonempty set
$A$ of segments and, in each selected segment, a nonempty set of milestones to query. Let $Q_A$ be the queried
milestones, $K'=|Q_A|$, and $h_j$ the number of unqueried milestones of segment $j\in A$ that precede its first queried
one.

\begin{theorem}[Cycle-aware certificate for unqueried prefixes]\label{thm:cyclecert}
Let $B\ge k$ and
\[
  D_B=\min\Bigl\{C_A(B),\ \frac{(B-k+K_A)^a}{a!}\prod_{j\in A}\frac1{h_j+1}\Bigr\}.
\]
\begin{enumerate}
\item[(i)] For every $p\in[0,1]^k$ and every process in $\Dcl(p)$,
$\Prob(\text{completion within }B)\le V_B(p)\le D_B\prod_{i\in Q_A}p_i$.
\item[(ii)] With $\tau'=-\log\bigl(1-(\epsilon/D_B)^{1/K'}\bigr)$, querying each $i\in Q_A$ exactly
$\lceil c_\alpha/(K'\tau')\rceil$ times and certifying if and only if all responses are zero is valid, certifies with
probability one when the queried probabilities are zero, and uses fewer than $c_\alpha/\tau'+K'$ responses.
\item[(iii)] As $B\to\infty$, the second term of $D_B$ divided by the first tends to $\prod_{j\in A}(h_j+1)^{-1}$. If
$h_j=0$ for every $j\in A$, the second term is never the smaller one, and if moreover every selected milestone is
queried, the rule is that of Theorem~\ref{thm:finite}.
\end{enumerate}
\end{theorem}

The gain arises only from unqueried prefixes of selected segments: with sure prefix steps every cycle of segment $j$
spends $h_j$ attempts before its first queried one, so $\kappa_j\ge h_j+1$ in Lemma~\ref{lem:renewal}, and the same
bound follows by monotonicity otherwise. Theorem~\ref{thm:cyclecert-general} allows known or capped success
probabilities below one.

The cycle-aware certificate is also asymptotically optimal when only a specified set $Q$ of milestones is queryable
and every other milestone has a known or capped success probability. Validity is required only over this restricted
class. The result assumes fixed positive declared constants and the phase-interior and rounding conditions of
Theorem~\ref{thm:restricted-full}. A segment's number of queryable milestones replaces its total length in the
active/free classification.

The leading constant involves the declared constants and a factor $1/\sigma_j$ for each active segment. Here
$\sigma_j$ is the expected number of attempts in one cycle of segment $j$ up to either a failure in the unqueried
prefix or the first attempt at a queryable milestone, including that terminating attempt, at the declared constants.
Optimal limiting allocation weights are uniform on the queryable milestones of active segments and zero elsewhere. For
$Q=\{1,\dots,k\}$ this recovers Theorems~\ref{thm:sharp}--\ref{thm:alloc}; otherwise the class and oracle differ,
so the result does not improve the constant of Theorem~\ref{thm:sharp} (Example~\ref{ex:freeq}).

\begin{corollary}[A six-transition family]\label{cor:six}
Consider one full-reset segment of six charged transitions with parameters $(1,q_1,q_2,q_3,1,1)$, where the first,
fifth and sixth transitions are known to succeed and $q\in[0,1]^3$ is queried. For $B\ge6$, $V_B\le D_Bq_1q_2q_3$ with
$D_B=\min\{B-5,B/2\}$, which is $B/2$ for $B\ge10$. Along sequences with $B_n\to\infty$, $B_np_n^3\to0$ and
$c_{\alpha_n}/t_n\to\infty$, and with $\epsilon_n=V_{B_n}(1,p_n,p_n,p_n,1,1)$, the cycle-aware certificate attains
$\inf\E_0N\sim c_{\alpha_n}(B_n/(2\epsilon_n))^{1/3}$ in the class with these sure transitions, whereas the window
certificate has $\tau/t\to2^{-1/3}$ and costs asymptotically $2^{1/3}\approx1.26$ times the optimum.
\end{corollary}

\section{Numerical examples and a conditional historical reanalysis}\label{sec:numerics}

The numbers below evaluate exact chain recursions in binary64 or in exact rationals, or come from an
integer-arithmetic branch and bound, as indicated. They illustrate the theorems and are not used in their proofs
(Appendix~\ref{app:numerics}).

\subsection{Evidence across budget regimes}

\begin{figure}[t]
  \centering
  \includegraphics[width=\columnwidth]{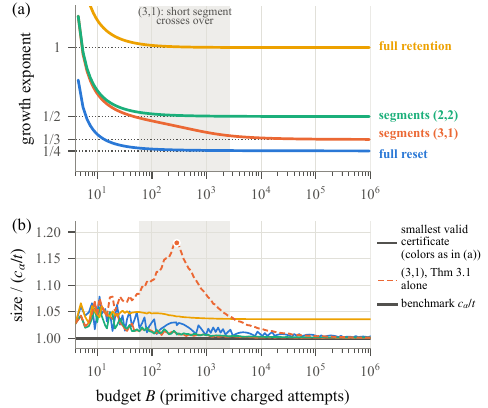}
  \caption{Zero-world evidence to certify $V_B\le10^{-4}$ at $\alpha=1/40$ with $k=4$ milestones; its minimum lies between
  the lower benchmark $c_\alpha/t$ and the smallest valid prospective certificate. (a) Growth exponent
  $\Delta\log(c_\alpha/t)/\Delta\log B$, cut off near $B=k$; dotted: rare-risk phase-interior exponents $a/K_A$ as $\epsilon\to0$ (Theorem~\ref{thm:sharp}), for reference
  alongside the fixed-$\epsilon$ curves. (b) Smallest
  valid certificate (Theorem~\ref{thm:finite} or Proposition~\ref{prop:crosscert}) over $c_\alpha/t$; dashed:
  Theorem~\ref{thm:finite} alone for $(3,1)$; the dot ($B=285$) marks where the smaller unrounded cost switches from
  both segments to the long one. Shaded: $1\le Bp_*\le10$ for $(3,1)$, where the short segment crosses over and the
  lower benchmark's growth exponent falls from $1/2$ to $1/3$.}
  \label{fig:complexity}
\end{figure}

All reset rules need the same evidence at $B=k$, where no failure can be absorbed; beyond it the growth exponents of
the lower benchmark, valid since $\I^*\le t$, approach $a/K_A$ (Figure~\ref{fig:complexity}(a)). At $B=10^6$, the
smallest displayed certificate uses about $1.2\times10^3$ responses under full reset and $1.6\times10^7$ under full
retention. Under full retention, the lower benchmark is within $k$ responses of the minimum
(Proposition~\ref{prop:schur}). The smallest certificate's excess of about $4\%$ over that benchmark therefore
reflects its own cost, with negligible uncertainty about the optimum. On the plotted budget grid, with Theorem~\ref{thm:finite} alone the
bracket is widest where the short segment of $(3,1)$ crosses over, $18\%$ at $B=279$ (Figure~\ref{fig:complexity}(b)),
close to the limiting factor $1.18$; the smaller of the two certificates keeps it below $3\%$ for $B\ge30$. At the
smallest budgets the window certificate is smaller.

\subsection{Cycle-aware accounting}

\begin{table}[t]
  \centering
  \caption{Six-transition family of Corollary~\ref{cor:six}, $\epsilon=1/100$, $\alpha=1/40$. $B$ counts primitive
  charged attempts, including the three sure transitions; responses are those at the queried milestones,
  $3\lceil c_\alpha/(3\tau)\rceil$ (window) and $3\lceil c_\alpha/(3\tau')\rceil$ (cycle-aware). The restricted
  diagonal threshold $p_\star$ lies in exact rational root brackets; other entries are binary64.}
  \label{tab:six}
  \small
  \setlength{\tabcolsep}{3.6pt}
  \begin{tabular}{rcccccc}
    \toprule
     & & & & & \multicolumn{2}{c}{responses}\\
    \cmidrule(lr){6-7}
    $B$ & $p_\star$ & $\tau/t$ & $\tau'/t$ & $c_\alpha/t$ & window & cycle\\
    \midrule
    100 & 0.0600927 & 0.78045 & 0.97234 & 59.52 & 78 & 63\\
    185 & 0.0485094 & 0.78238 & 0.98159 & 74.18 & 96 & 78\\
    \bottomrule
  \end{tabular}
\end{table}

For the six-transition family at $\epsilon=1/100$ (Table~\ref{tab:six}), the cycle-aware certificate uses 63 zero
responses instead of 78 at $B=100$ and 78 instead of 96 at $B=185$, about $19\%$ fewer. Its counts are $5.8\%$ and
$5.1\%$ above the lower benchmark $c_\alpha/t$, and the window counts are $31\%$ and $29\%$ above it. These operating
points are not on phase-interior sequences ($Bp_\star^3\approx0.022$ and $0.021$), and the ratios compare cost
benchmarks rather than gaps to a finite-budget minimax value. Figure~\ref{fig:cycle} (Appendix~\ref{app:numerics})
extends this comparison to other budgets.

\subsection{A conditional reanalysis of a published clean sheet}\label{sec:gdm}

\citet{phuong2024evaluating} applied the milestone method to self-proliferation tasks: they ran the agent ten times
from a prepared state at the start of each milestone and reported the upper end of a one-sided $97.5\%$ Bayesian
credible interval for the probability that one complete trial succeeds, with independent uniform priors (their
Appendix~E.4). For the four-milestone task ``Implement paper'', both models in their Table~11 (Gemini 1.0 Pro and Ultra)
failed all ten unguided runs at every milestone, and the reported bound is below one in 2{,}700. A credible bound is a
posterior quantile; unlike a confidence statement, and unlike our certificates, it does not bound the error
probability of a procedure at each fixed parameter. \citeauthor{phuong2024evaluating} state no guarantee for a campaign
of attempts, and we do not compare their bound with ours.

\begin{table}[t]
  \centering
  \caption{Conditional reanalysis of a four-milestone record with ten zero responses per milestone
  ($\alpha=1/40$, $\epsilon=1/100$) under the model of Section~\ref{sec:model};
  $B_{\max}$ counts primitive charged attempts, or complete trials where stated.}
  \label{tab:gdm}
  \footnotesize
  \setlength{\tabcolsep}{2.8pt}
  \begin{tabular}{@{}lrl@{}}
    \toprule
    Reset rule (unit) & $B_{\max}$ & Statement\\
    \midrule
    Full reset & 169 & Analytic upper-bound certificate\\
    Full reset & 185 & Computer-assisted exact-risk certificate\\
    Full retention & 10 & Exact full-retention result\\
    Full reset (trials) & 166 & Exact complete-trial result\\
    \bottomrule
  \end{tabular}
\end{table}

We use the record only as input to a conditional, frequentist calculation: suppose that the forty unguided responses
of either model satisfy Assumption~\ref{ass:O} with counts $n=(10,10,10,10)$ fixed in advance, and that the deployment
process lies in $\Dcl(p)$ for the same $p$. Proposition~\ref{prop:plausibility} with $\alpha=1/40$ and $\epsilon=1/100$
then certifies the attempt budgets of Table~\ref{tab:gdm}; the complete-trial row follows from the same plausibility
argument applied to the risk of independent complete trials. On $\Pi=\{u:\sum_iu_i\le\log(40)/10\}$ the product $\prod_ip_i$ is
maximal at the symmetric point $p_i=\bar p=1-40^{-1/40}\approx0.0881$, by concavity of $\log(1-e^{-u})$.

Under full reset, Theorem~\ref{thm:finite} gives $\sup_{u\in\Pi}F_B(u)\le(B-3)\bar p^4$, and exact rational arithmetic gives
$166\,\bar p^4<1/100<167\,\bar p^4$, so this analytic certificate covers $B\le169$. A computer-assisted proof, an
integer-arithmetic branch and bound over $\Pi$, shows $F_{185}(u)<1/100$ throughout $\Pi$. At $B=186$, the symmetric
point already has $V_{186}\approx0.0100093$, so 185 is the largest budget the plausibility rule can certify.
The computation bounds the worst-case risk at $B=185$; it does not locate the maximizing parameter or the exact
margin below $1/100$ (Appendix~\ref{app:numerics}).

A rational probability vector $q$ with $-\log(\one-q)\in\Pi$ and a smaller first coordinate has $V_{185}(q)\approx0.0099791$, above the symmetric value $V_{185}(\bar p\one)\approx0.0099549$, so the maximizer is asymmetric. The step from
169 to 185 is consistent with the cost of failed cycles, since a completion window can start only where a cycle
starts; at the symmetric point cycles have mean length $1+\bar p+\bar p^2+\bar p^3\approx1.097$
(Lemma~\ref{lem:renewal}), close to $185/169\approx1.095$. This is a heuristic, not a proof.

Under full retention,
Proposition~\ref{prop:schur} places $\sup_{u\in\Pi}F_B(u)$ at the symmetric point, where exact arithmetic gives
$V_{10}<1/100<V_{11}$. In complete trials the risk $1-(1-\prod_ip_i)^n$ is maximal at the symmetric point, and exact
arithmetic gives 166 trials safe and 167 unsafe.

Table~\ref{tab:sensitivity} varies $k$, $n$, $\epsilon$ and
$\alpha$: full reset certifies more attempts in every case, by factors from 2 to about 7{,}200.

The calculation inherits every assumption of the model. Responses from prepared milestone states need not represent
the states an agent reaches by itself, and \citet{hojmark2024analyzing} report that milestone-based estimates can
understate end-to-end success rates; both bear on the transport assumption. The four milestones need
not be the only route to the goal, and attempts may share a latent cause.

\section{Related work, limitations and conclusion}\label{sec:related}

\paragraph{Milestone evaluations and rare events.} \citet{phuong2024evaluating} introduced Bayesian milestone
bounds on end-to-end success and expert-help scores in bits. \citet{hojmark2024analyzing} analyze these methods as
probability estimators, including their bias and variance. Estimating a
rare probability as a product of conditional probabilities from intermediate states is the idea of multilevel
splitting \citep{kahn1951estimation,glasserman1999multilevel,au2001estimation,cerou2007adaptive}; our oracle is of this
kind, but we certify a campaign-level probability.

\paragraph{Zero-failure and component testing.} One-sided bounds from zero failures are classical
\citep{clopper1934use,hanley1983nothing}, and demonstrating very small failure probabilities by testing needs very
many trials \citep{littlewood1993validation}. \citet{buehler1957confidence} bounds the product of two binomial parameters.
\citet{bishop2020conservative} bound a system's probability of failure on demand from failure-free component tests
without assuming independent failures. For positive $N_{\min}$, their conservative upper bound is
$\min\{1,\log(1/\alpha)/N_{\min}\}$, where $N_{\min}$ is the smallest total number of component tests over a minimal
cut set. A system that fails only when every component fails gains nothing over black-box testing when failures may
have arbitrary dependence. Passing all milestones in one traversal has this logical structure, with component failure corresponding
to milestone success. Our multiplicative gain requires (D3), which rules out this dependence
(Example~\ref{ex:average}). \citet{bishop2022optimising} optimize the allocation by linear programming.

\paragraph{Certifying risk of language models.} \citet{wang2026catastrophic} bound the probability that an $n$-turn
conversation drawn from a specified Markov distribution over a query graph yields a catastrophic response, with
binomial confidence intervals on sampled conversations. \citet{davidov2026calibrated} give conformal lower predictive
bounds on the number of generations until the first unsafe one, from censored generation streams, and optimize the allocation of an
expected calibration budget across prompts; with one milestone our campaign risk is the distribution function of such a time. \citet{cheng2026sharp} derive matching bounds, up to constants, on expected interval width for benchmark scores under a hard budget of repeated runs.
There each observation is a complete trajectory or generation, and the budget is the evaluator's; here observations
are single attempts from prepared states, and budget, resets and adaptivity belong to the certified agent.

\paragraph{Sample complexity and verification.} Lower bounds based on a change of measure and the allocations that attain them go
back to \citet{chernoff1959sequential} and are developed for best-arm identification by \citet{kaufmann2016complexity}
and \citet{garivier2016optimal}; here the chain enters through the geometry of the unsafe set. Statistical model
checking tests bounded-horizon reachability from sampled complete runs \citep{younes2002probabilistic,legay2010statistical},
and \citet{ashok2019pac} propagate per-transition confidence bounds. \citet{liu2026certified} compose lower bounds on per-turn safety
that hold in every safe dialogue state over a fixed conversation horizon, without our milestone reset semantics.

\paragraph{Renewal theory and restarts.} Waiting times for success runs
\citep{feller1968introduction,philippou1982waiting,dilworth2015success,dilworth2016applications} and completion times of
jobs that restart or resume after failures \citep{kulkarni1987completion,asmussen2008asymptotic} are classical.
\citet{emadi2026critical} relates the samples needed to attribute outcomes to the longest uninspected stretch of a
multi-stage process, where inspections do not change the dynamics; attack-graph models with adaptive attackers
\citep{ling2026security,tan2026security} optimize against known success probabilities.

\paragraph{Limitations.} The transport assumption is not tested; it can fail when prepared milestone states differ
from those reached in deployment, when attempts share a latent cause, or when the agent improves across attempts.
Condition (D1) excludes alternative routes and skipped milestones, and reaching milestone states is not costed. The
complexity results concern the zero world only; leading constants hold in phase interiors and at single crossovers
with fixed $\lambda$, allocation results concern ratios of expected counts in the limit, and the restricted results assume
fixed declared constants. The certificate at $B=185$ is computer-assisted and not formally verified.

\paragraph{Conclusion.} A clean sheet bounds campaign risk only through a model of what failures cost. Under the
reset-segment model and domination conditions, counting completion windows gives a certificate for any specified
budget. In rare-risk phase interiors, it attains the optimal leading zero-world cost, and every asymptotically
optimal audit must divide its expected response count equally among active milestones in the limit, with a vanishing share on free ones. A crossover-aware
certificate attains the optimal leading constant at a single crossover. Crossovers of several segments, fixed risk, finite-budget
designs and nonzero safe parameters remain open.

\section*{Impact Statement}

This paper studies statistical certificates that bound the probability that an AI agent completes a multi-step task
within a budget of attempts. Such bounds may inform deployment decisions about agents with dangerous capabilities.
Their main risk is misuse as unconditional safety claims: the certificates hold only under stated modeling and
transport assumptions that audits do not test, and a certified budget can change substantially when those
assumptions change. We state these assumptions next to the results and report the reanalysis of a published
evaluation as a conditional calculation, not as a statement about the evaluated systems.

\bibliography{references}

\begin{thebibliography}{43}
\providecommand{\natexlab}[1]{#1}
\providecommand{\url}[1]{\texttt{#1}}
\expandafter\ifx\csname urlstyle\endcsname\relax
  \providecommand{\doi}[1]{doi: #1}\else
  \providecommand{\doi}{doi: \begingroup \urlstyle{rm}\Url}\fi

\bibitem[Ashok et~al.(2019)Ashok, K{\v{r}}et{\'\i}nsk{\'y}, and
  Weininger]{ashok2019pac}
Ashok, P., K{\v{r}}et{\'\i}nsk{\'y}, J., and Weininger, M.
\newblock {PAC} statistical model checking for {Markov} decision processes and
  stochastic games.
\newblock In \emph{Computer Aided Verification (CAV)}, volume 11561 of
  \emph{Lecture Notes in Computer Science}, pp.\  497--519. Springer, 2019.

\bibitem[Asmussen et~al.(2008)Asmussen, Fiorini, Lipsky, Rolski, and
  Sheahan]{asmussen2008asymptotic}
Asmussen, S., Fiorini, P., Lipsky, L., Rolski, T., and Sheahan, R.
\newblock Asymptotic behavior of total times for jobs that must start over if a
  failure occurs.
\newblock \emph{Mathematics of Operations Research}, 33\penalty0 (4):\penalty0
  932--944, 2008.

\bibitem[Au \& Beck(2001)Au and Beck]{au2001estimation}
Au, S.-K. and Beck, J.~L.
\newblock Estimation of small failure probabilities in high dimensions by
  subset simulation.
\newblock \emph{Probabilistic Engineering Mechanics}, 16\penalty0 (4):\penalty0
  263--277, 2001.

\bibitem[Bishop \& Povyakalo(2020)Bishop and Povyakalo]{bishop2020conservative}
Bishop, P. and Povyakalo, A.
\newblock A conservative confidence bound for the probability of failure on
  demand of a software-based system based on failure-free tests of its
  components.
\newblock \emph{Reliability Engineering \& System Safety}, 203:\penalty0
  107060, 2020.

\bibitem[Bishop \& Povyakalo(2022)Bishop and Povyakalo]{bishop2022optimising}
Bishop, P. and Povyakalo, A.
\newblock Optimising the reliability that can be claimed for a software-based
  system based on failure-free tests of its components.
\newblock In \emph{Computer Safety, Reliability, and Security (SAFECOMP 2022)},
  volume 13414 of \emph{Lecture Notes in Computer Science}, pp.\  365--378.
  Springer, 2022.

\bibitem[Boland et~al.(1994)Boland, El-Neweihi, and Proschan]{boland1994schur}
Boland, P.~J., El-Neweihi, E., and Proschan, F.
\newblock Schur properties of convolutions of exponential and geometric random
  variables.
\newblock \emph{Journal of Multivariate Analysis}, 48\penalty0 (1):\penalty0
  157--167, 1994.

\bibitem[Brown(1990)]{brown1990error}
Brown, M.
\newblock Error bounds for exponential approximations of geometric
  convolutions.
\newblock \emph{Annals of Probability}, 18\penalty0 (3):\penalty0 1388--1402,
  1990.

\bibitem[Buehler(1957)]{buehler1957confidence}
Buehler, R.~J.
\newblock Confidence intervals for the product of two binomial parameters.
\newblock \emph{Journal of the American Statistical Association}, 52\penalty0
  (280):\penalty0 482--493, 1957.

\bibitem[C{\'e}rou \& Guyader(2007)C{\'e}rou and Guyader]{cerou2007adaptive}
C{\'e}rou, F. and Guyader, A.
\newblock Adaptive multilevel splitting for rare event analysis.
\newblock \emph{Stochastic Analysis and Applications}, 25\penalty0
  (2):\penalty0 417--443, 2007.

\bibitem[Cheng et~al.(2026)Cheng, Du, Liu, Chen, Lyu, Wei, Zeng, and
  Lin]{cheng2026sharp}
Cheng, Y., Du, R., Liu, Z., Chen, Q., Lyu, H., Wei, Y., Zeng, Y., and Lin, B.
\newblock Sharp limits for honest uncertainty in hard-budget repeated
  evaluation.
\newblock \emph{arXiv preprint arXiv:2609.29140}, 2026.

\bibitem[Chernoff(1959)]{chernoff1959sequential}
Chernoff, H.
\newblock Sequential design of experiments.
\newblock \emph{Annals of Mathematical Statistics}, 30\penalty0 (3):\penalty0
  755--770, 1959.

\bibitem[Clopper \& Pearson(1934)Clopper and Pearson]{clopper1934use}
Clopper, C.~J. and Pearson, E.~S.
\newblock The use of confidence or fiducial limits illustrated in the case of
  the binomial.
\newblock \emph{Biometrika}, 26\penalty0 (4):\penalty0 404--413, 1934.

\bibitem[Davidov et~al.(2026)Davidov, Feldman, Freidkin, and
  Romano]{davidov2026calibrated}
Davidov, H., Feldman, S., Freidkin, G., and Romano, Y.
\newblock Calibrated predictive lower bounds on time-to-unsafe-sampling in
  {LLMs}.
\newblock In \emph{Proceedings of the 29th International Conference on
  Artificial Intelligence and Statistics}, volume 300 of \emph{Proceedings of
  Machine Learning Research}, pp.\  5032--5040, 2026.

\bibitem[Dilworth \& Mane(2015)Dilworth and Mane]{dilworth2015success}
Dilworth, S.~J. and Mane, S.~R.
\newblock Success run waiting times and {Fuss}--{Catalan} numbers.
\newblock \emph{Journal of Probability and Statistics}, 2015:\penalty0 482462,
  2015.

\bibitem[Dilworth \& Mane(2016)Dilworth and Mane]{dilworth2016applications}
Dilworth, S.~J. and Mane, S.~R.
\newblock Applications of {Fuss}--{Catalan} numbers to success runs of
  {Bernoulli} trials.
\newblock \emph{Journal of Probability and Statistics}, 2016:\penalty0 2071582,
  2016.

\bibitem[Emadi(2026)]{emadi2026critical}
Emadi, S.~M.
\newblock The critical horizon: Inspection design principles for multi-stage
  operations and deep reasoning.
\newblock \emph{arXiv preprint arXiv:2602.09394}, 2026.

\bibitem[Feller(1968)]{feller1968introduction}
Feller, W.
\newblock \emph{An Introduction to Probability Theory and Its Applications},
  volume~I.
\newblock Wiley, third edition, 1968.

\bibitem[Garivier \& Kaufmann(2016)Garivier and Kaufmann]{garivier2016optimal}
Garivier, A. and Kaufmann, E.
\newblock Optimal best arm identification with fixed confidence.
\newblock In \emph{Proceedings of the 29th Annual Conference on Learning Theory
  (COLT)}, volume~49 of \emph{Proceedings of Machine Learning Research}, pp.\
  998--1027, 2016.

\bibitem[Glasserman et~al.(1999)Glasserman, Heidelberger, Shahabuddin, and
  Zajic]{glasserman1999multilevel}
Glasserman, P., Heidelberger, P., Shahabuddin, P., and Zajic, T.
\newblock Multilevel splitting for estimating rare event probabilities.
\newblock \emph{Operations Research}, 47\penalty0 (4):\penalty0 585--600, 1999.

\bibitem[Hanley \& Lippman-Hand(1983)Hanley and
  Lippman-Hand]{hanley1983nothing}
Hanley, J.~A. and Lippman-Hand, A.
\newblock If nothing goes wrong, is everything all right? {Interpreting} zero
  numerators.
\newblock \emph{Journal of the American Medical Association}, 249\penalty0
  (13):\penalty0 1743--1745, 1983.

\bibitem[H{\o}jmark et~al.(2024)H{\o}jmark, Pimpale, Panickssery, Hobbhahn, and
  Scheurer]{hojmark2024analyzing}
H{\o}jmark, A., Pimpale, G., Panickssery, A., Hobbhahn, M., and Scheurer, J.
\newblock Analyzing probabilistic methods for evaluating agent capabilities.
\newblock In \emph{Socially Responsible Language Modelling Research (SoLaR)
  Workshop at NeurIPS}, 2024.
\newblock arXiv:2409.16125.

\bibitem[Iyengar(2005)]{iyengar2005robust}
Iyengar, G.~N.
\newblock Robust dynamic programming.
\newblock \emph{Mathematics of Operations Research}, 30\penalty0 (2):\penalty0
  257--280, 2005.

\bibitem[Kahn \& Harris(1951)Kahn and Harris]{kahn1951estimation}
Kahn, H. and Harris, T.~E.
\newblock Estimation of particle transmission by random sampling.
\newblock In \emph{Monte Carlo Method}, volume~12 of \emph{National Bureau of
  Standards Applied Mathematics Series}, pp.\  27--30. United States Government
  Printing Office, 1951.

\bibitem[Kalashnikov(1997)]{kalashnikov1997geometric}
Kalashnikov, V.
\newblock \emph{Geometric Sums: Bounds for Rare Events with Applications. Risk
  Analysis, Reliability, Queueing}.
\newblock Kluwer, 1997.

\bibitem[Kaufmann et~al.(2016)Kaufmann, Capp{\'e}, and
  Garivier]{kaufmann2016complexity}
Kaufmann, E., Capp{\'e}, O., and Garivier, A.
\newblock On the complexity of best-arm identification in multi-armed bandit
  models.
\newblock \emph{Journal of Machine Learning Research}, 17\penalty0
  (1):\penalty0 1--42, 2016.

\bibitem[Keilson(1979)]{keilson1979markov}
Keilson, J.
\newblock \emph{Markov Chain Models -- Rarity and Exponentiality}.
\newblock Springer, 1979.

\bibitem[Kulkarni et~al.(1987)Kulkarni, Nicola, and
  Trivedi]{kulkarni1987completion}
Kulkarni, V.~G., Nicola, V.~F., and Trivedi, K.~S.
\newblock The completion time of a job on multimode systems.
\newblock \emph{Advances in Applied Probability}, 19\penalty0 (4):\penalty0
  932--954, 1987.

\bibitem[Legay et~al.(2010)Legay, Delahaye, and Bensalem]{legay2010statistical}
Legay, A., Delahaye, B., and Bensalem, S.
\newblock Statistical model checking: An overview.
\newblock In \emph{Runtime Verification (RV)}, volume 6418 of \emph{Lecture
  Notes in Computer Science}, pp.\  122--135. Springer, 2010.

\bibitem[Ling et~al.(2026)Ling, {\v{C}}ern{\'y}, Chin, Iyengar, and
  Kroer]{ling2026security}
Ling, C.~K., {\v{C}}ern{\'y}, J., Chin, H.~H., Iyengar, G., and Kroer, C.
\newblock Security games with layered defenses: Adaptive adversaries and
  {Gittins} indices.
\newblock In \emph{Proceedings of the AAAI Conference on Artificial
  Intelligence}, volume~40, pp.\  17120--17128, 2026.
\newblock \doi{10.1609/aaai.v40i20.38761}.

\bibitem[Littlewood \& Strigini(1993)Littlewood and
  Strigini]{littlewood1993validation}
Littlewood, B. and Strigini, L.
\newblock Validation of ultrahigh dependability for software-based systems.
\newblock \emph{Communications of the ACM}, 36\penalty0 (11):\penalty0 69--80,
  1993.

\bibitem[Liu et~al.(2026)Liu, Chong, Yang, Zhang, Chen,
  et~al.]{liu2026certified}
Liu, Y., Chong, B., Yang, W., Zhang, S., Chen, Y., et~al.
\newblock Certified multi-turn robustness for {LLM} safety via compositional
  bounds and safety persistence.
\newblock \emph{arXiv preprint arXiv:2608.20820}, 2026.

\bibitem[Lorden(1970)]{lorden1970excess}
Lorden, G.
\newblock On excess over the boundary.
\newblock \emph{Annals of Mathematical Statistics}, 41\penalty0 (2):\penalty0
  520--527, 1970.

\bibitem[Mao et~al.(2010)Mao, Hu, and Zhao]{mao2010ordering}
Mao, T., Hu, T., and Zhao, P.
\newblock Ordering convolutions of heterogeneous exponential and geometric
  distributions revisited.
\newblock \emph{Probability in the Engineering and Informational Sciences},
  24\penalty0 (3):\penalty0 329--348, 2010.

\bibitem[Nilim \& El~Ghaoui(2005)Nilim and El~Ghaoui]{nilim2005robust}
Nilim, A. and El~Ghaoui, L.
\newblock Robust control of {Markov} decision processes with uncertain
  transition matrices.
\newblock \emph{Operations Research}, 53\penalty0 (5):\penalty0 780--798, 2005.

\bibitem[Philippou \& Muwafi(1982)Philippou and Muwafi]{philippou1982waiting}
Philippou, A.~N. and Muwafi, A.~A.
\newblock Waiting for the $k$-th consecutive success and the {Fibonacci}
  sequence of order $k$.
\newblock \emph{Fibonacci Quarterly}, 20\penalty0 (1):\penalty0 28--32, 1982.

\bibitem[Phuong et~al.(2024)Phuong, Aitchison, Catt, Cogan, Kaskasoli,
  et~al.]{phuong2024evaluating}
Phuong, M., Aitchison, M., Catt, E., Cogan, S., Kaskasoli, A., et~al.
\newblock Evaluating frontier models for dangerous capabilities.
\newblock \emph{arXiv preprint arXiv:2403.13793}, 2024.

\bibitem[Pr{\'e}kopa(1973)]{prekopa1973logarithmic}
Pr{\'e}kopa, A.
\newblock On logarithmic concave measures and functions.
\newblock \emph{Acta Scientiarum Mathematicarum (Szeged)}, 34:\penalty0
  335--343, 1973.

\bibitem[R{\'e}nyi(1956)]{renyi1956characterization}
R{\'e}nyi, A.
\newblock A characterization of {Poisson} processes.
\newblock \emph{A Magyar Tudom{\'a}nyos Akad{\'e}mia Matematikai Kutat{\'o}
  Int{\'e}zet{\'e}nek K{\"o}zlem{\'e}nyei}, 1\penalty0 (4):\penalty0 519--527,
  1956.
\newblock In Hungarian, with English and Russian summaries. Volume dated 1956;
  published 1957.

\bibitem[Shevlane et~al.(2023)Shevlane, Farquhar, Garfinkel, Phuong,
  Whittlestone, et~al.]{shevlane2023model}
Shevlane, T., Farquhar, S., Garfinkel, B., Phuong, M., Whittlestone, J., et~al.
\newblock Model evaluation for extreme risks.
\newblock \emph{arXiv preprint arXiv:2305.15324}, 2023.

\bibitem[Tan et~al.(2026)Tan, Chin, and Ling]{tan2026security}
Tan, R. K.~M., Chin, H.~H., and Ling, C.~K.
\newblock Security games on series--parallel attack graphs with adaptive
  attackers.
\newblock \emph{arXiv preprint arXiv:2608.21259}, 2026.

\bibitem[Wald(1944)]{wald1944cumulative}
Wald, A.
\newblock On cumulative sums of random variables.
\newblock \emph{Annals of Mathematical Statistics}, 15\penalty0 (3):\penalty0
  283--296, 1944.

\bibitem[Wang et~al.(2026)Wang, Chaudhary, Hu, Ruan, Gupta, and
  Singh]{wang2026catastrophic}
Wang, C., Chaudhary, I., Hu, Q., Ruan, W., Gupta, R., and Singh, G.
\newblock How catastrophic is your {LLM}? {Certifying} risks in conversation.
\newblock In \emph{International Conference on Learning Representations
  (ICLR)}, 2026.
\newblock arXiv:2510.03969.

\bibitem[Younes \& Simmons(2002)Younes and Simmons]{younes2002probabilistic}
Younes, H. L.~S. and Simmons, R.~G.
\newblock Probabilistic verification of discrete event systems using acceptance
  sampling.
\newblock In \emph{Computer Aided Verification (CAV)}, volume 2404 of
  \emph{Lecture Notes in Computer Science}, pp.\  223--235. Springer, 2002.

\end{thebibliography}
\bibliographystyle{icml2026}

\newpage
\appendix
\onecolumn
\section{Notation}\label{app:notation}

\begin{center}
\small
\begin{tabular}{@{}p{0.27\textwidth}p{0.68\textwidth}@{}}
\toprule
Symbol & Meaning\\
\midrule
$k$, $s$, $\ell_j$, $L$, $E_j$ & milestones, segments, segment lengths, $L=\max_j\ell_j$, milestones of segment $j$\\
$r(i)$ & state after a failed attempt from state $i$ (start of the segment of milestone $i+1$)\\
$p$, $\beta_j$, $\pi_{j,h}$, $\kappa_j$ & milestone probabilities, segment success probability $\prod_{E_j}p_i$, product
of the first $h$ probabilities of segment $j$, mean cycle length $\sum_{h<\ell_j}\pi_{j,h}$\\
$B$, $V_B(p)$, $W^p_R(i)$ & budget in charged attempts; completion probability within $B$ from state $0$; the same
from state $i$ with $R$ attempts left\\
$\Dcl(p)$ & deployment processes satisfying (D1)--(D3)\\
$N_i$, $N$, $\E_0$ & responses at milestone $i$, total responses, expectation in the zero world $p=0$\\
$c_\alpha$, $u$, $F_B$ & $\log(1/\alpha)$; $u_i=-\log(1-p_i)$; $F_B(u)=V_B(\one-e^{-u})$\\
$U$, $S$, $I_w$, $\I^*$, $t$, $p_*$ & strictly unsafe set, its closure, information of allocation $w$, its maximum,
diagonal threshold, $p_*=1-e^{-t}$\\
$A$, $a$, $E_A$, $K_A$ & selected segments, their number, their milestones, $K_A=|E_A|$\\
$C_A(B)$, $r_A$, $\tau_A$, $w_A$ & $\binom{B-k+a}{a}$, $(\epsilon/C_A(B))^{1/K_A}$, $-\log(1-r_A)$, uniform weights on
$E_A$\\
$(B_n,p_n)$, $\epsilon_n$, $t_n$ & phase-interior sequence, its risk level $V_{B_n}(p_n\one)$, $-\log(1-p_n)$\\
$Q$, $c_i$, $\tilde c_i$, $\nu_j$ & queryable milestones, declared constants, $\tilde c_i=c_i$ off $Q$ and $1$ on $Q$,
$\nu_j=|Q\cap E_j|$\\
$Q_A$, $K'$, $\Hc_j$, $h_j$ & queried milestones of a cycle-aware design, their number, unqueried prefix of segment
$j$ and its length\\
$\sigma_j$, $\rho_j$, $\bar c_j$ & mean truncated cycle length, $\prod_{\Hc_j}\tilde c_i/\sigma_j$, product of $\tilde c_i$ after
the first queried milestone\\
$D_{A,Q_A}(B)$, $r'$, $\tau'$ & cycle-aware coefficient, $(\epsilon/D_{A,Q_A}(B))^{1/K'}$, $-\log(1-r')$\\
$F^Q_B$, $\bar\epsilon_Q$, $t_Q$, $I^Q_w$, $\I^Q_*$ & restricted objects of Appendix~\ref{app:classes}\\
$\Gamma_A$, $K'_A$ & $\prod_{j\in A}\rho_j\bar c_j$ with $Q_A=Q\cap E_A$, and $|Q\cap E_A|$\\
$M$, $m$, $E_M$, $\lambda$ & crossover segment, its length and milestones; $\lambda=\lim B_np_n^m$\\
$h_a$, $e_a$, $\omega(\lambda)$ & crossover profile $\E(1-E/\lambda)_+^a$, its elasticity, limiting allocation of
Theorem~\ref{thm:crossover}\\
$B_\times$, $\tau_\times$ & $B-k+a+1$; root of $(1-e^{-\tau})^{K_A}(B_\times^a/a!)h_a(B_\times\tau^m)=\epsilon$ (Proposition~\ref{prop:crosscert})\\
\bottomrule
\end{tabular}
\end{center}

\section{Proofs for Section~\ref{sec:model}}\label{app:model}

Let $W^p_R(i)$ be the probability that the independent chain reaches state $k$ within $R$ attempts from state $i$, so
that $V_B(p)=W^p_B(0)$.

\begin{lemma}[Monotone envelope]\label{lem:envelope}
\begin{enumerate}
\item[(a)] $W^p_R(i)$ is nondecreasing in $R$, in the state $i$, and coordinatewise in $p\in[0,1]^k$.
\item[(b)] If $B\ge k$ and $V_B(p)<1$, then $\lambda\mapsto V_B(p+\lambda(\one-p))$ is strictly increasing on
$[0,1]$. If $B<k$, then $V_B\equiv0$.
\item[(c)] Every process in $\Dcl(p)$ completes within $B$ attempts with probability at most $V_B(p)$. The
independent chain belongs to $\Dcl(p)$ and attains the bound.
\end{enumerate}
Lemma~\ref{lem:domination} consists of parts (a) and (c).
\end{lemma}

\begin{proof}
Monotonicity in the horizon follows from the inclusion of completion events. For $i<j$, every successful path from $i$
first hits $j$, because successes advance one state and failures never move forward; with first-passage time
$\theta_j$, the Markov property and horizon monotonicity give
$W^p_R(i)=\E^p_i[\one\{\theta_j\le R\}W^p_{R-\theta_j}(j)]\le W^p_R(j)$. For $p'\ge p$ we induct on $R$. At $R=0$
both sides equal $\one\{i=k\}$, and for $i<k$
\begin{align*}
W^{p'}_R(i)&=p'_{i+1}W^{p'}_{R-1}(i+1)+(1-p'_{i+1})W^{p'}_{R-1}(r(i))\\
&\ge p'_{i+1}W^{p}_{R-1}(i+1)+(1-p'_{i+1})W^{p}_{R-1}(r(i))\ge W^p_R(i),
\end{align*}
the last step because $W^p_{R-1}(i+1)\ge W^p_{R-1}(r(i))$ by the monotonicity in the state. For (b), the value along
the ray is a polynomial in $\lambda$ (a finite sum over paths), nondecreasing by (a), equal to one at $\lambda=1$
because $B\ge k$, and smaller at $\lambda=0$ by hypothesis. If it took equal values at two points it would be constant
on an interval, hence everywhere, a contradiction. For (c), induct on the remaining budget uniformly over histories.
At a history $h$ in state $i<k$ with next success probability $a(h)\le p_{i+1}$, the induction hypothesis bounds the
conditional completion probability by $a(h)W^p_{R-1}(i+1)+(1-a(h))W^p_{R-1}(r(i))\le W^p_R(i)$, using the
monotonicity in the state again; stopping gives zero. The base cases are immediate.
\end{proof}

\begin{example}[Average caps do not suffice]\label{ex:average}
Take $k=2$ in two segments of length one, budget $B=2$, and a campaign-level latent variable
$Z\sim\mathrm{Bernoulli}(r)$, $0<r<1$, such that every attempt succeeds if and only if $Z=1$. An attempt at either
milestone from a fresh campaign, for instance from a scripted state, succeeds with probability $r$, so the average
success rate at each milestone is $r$. The completion probability is $r$, whereas $V_2(r,r)=r^2$. Condition (D3) fails:
given any history in state $1$, the next attempt succeeds with probability one.
\end{example}

\begin{proof}[Proof of Lemma~\ref{lem:closure}]
For finite $u$, the point $u+\delta\one$ corresponds to $p+\lambda(\one-p)$ with $\lambda=1-e^{-\delta}$. At a point
with $F_B(u)=\epsilon<1$, Lemma~\ref{lem:envelope}(b) makes every such point with $\delta>0$ strictly unsafe, so
$S\subseteq\overline U$; continuity gives the reverse inclusion. Applied at $p=0$, the same lemma makes $x\mapsto
F_B(x\one)$ continuous and strictly increasing from $0$ to $1$, so $t$ exists and is unique. Each $I_w$ is an infimum
of linear functions of $w$, hence upper semicontinuous on the compact simplex, so the maximum is attained; and
$t\one\in S$ gives $I_w\le t$ for every $w$.
\end{proof}

Strict unsafety ($U$) is used for validity and the closure ($S$) only for information infima. If $B<k$ then
$V_B\equiv0$ and every parameter is safe; this trivial case is excluded throughout.

\begin{proposition}[Nonconvex unsafe sets]\label{prop:nonconvex}
Take full reset with $k=5$ and $B=9$. In the logit coordinates $\xi_i=\log\bigl(p_i/(1-p_i)\bigr)$ write
$\tilde F_9(\xi)=V_9(p)$; then $u_i=\log(1+e^{\xi_i})$ and $\tilde F_9(\xi)=F_9(u)$.
\begin{enumerate}
\item[(a)] Let $p_1=199/200$, $p_2=p_3=1-10^{-8}$ and $p_4=1-10^{-5}$, and let $p_5$ be the number with
$V_9(p)=1/100$; it is rational, and $p_5\approx0.0100000010$. Then $\log\tilde F_9$ is not concave on any
neighborhood of the corresponding point $\xi$, and neither $\{\tilde F_9\ge1/100\}$ nor $\{\tilde F_9>1/100\}$ is
convex.
\item[(b)] Let $p^\circ=(1-10^{-3},\,1-10^{-12},\,1-10^{-12},\,1-10^{-7},\,1/50)$ and
$\epsilon^\circ=V_9(p^\circ)\approx0.0199999999960$. Then $\log F_9$ is not concave on any neighborhood of the
corresponding point $u^\circ$, and for $\epsilon=\epsilon^\circ$ neither $S$ nor $U$ is convex.
\end{enumerate}
\end{proposition}

\begin{proof}
We first obtain an exact expression for the risk. We then show how positive curvature along a tangent to a risk
level set gives a nonconvex unsafe set. Finally, we verify that curvature at the two stated points in their
respective coordinates.

A failure at milestone~$5$ occurs at attempt $5$ or later and returns the chain to state $0$, after which five more
attempts are needed. Within nine attempts milestone~$5$ is therefore attempted at most once, and $V_9(p)=p_5C$, where
$C$ is the probability of reaching state $4$ within eight attempts. Before its final run of four successes a path
makes failed cycles; a cycle that fails at milestone $h\le4$ has length $h$ and probability
$f_h=p_1\cdots p_{h-1}(1-p_h)$, and the failed cycles have total length at most $4$. Hence
\[
  C=p_1p_2p_3p_4\sum_{j=0}^{4}W_j,\qquad W_0=1,\quad W_j=\sum_{h=1}^{j}f_hW_{j-h}.
\]
In (a), $p_5=1/(100\,C)$ with $C$ evaluated at $(p_1,\dots,p_4)$; this is rational and lies in $(0,1)$ because
$1/100<C<1$.

Now let $G$ be twice continuously differentiable near an interior point $x$ of its domain, with $G(x)=\epsilon$ and
$\partial_5G(x)>0$, and let
$v$ satisfy $\nabla G(x)^\top v=0$ and $c=v^\top\nabla^2G(x)\,v>0$. Fix $0<\delta<c/(2\,\partial_5G(x))$. As
$s\downarrow0$, Taylor's theorem gives
\[
  G(x\pm sv-\delta s^2e_5)=\epsilon+s^2\bigl(c/2-\delta\,\partial_5G(x)\bigr)+o(s^2),\qquad
  G(x-\delta s^2e_5)=\epsilon-\delta s^2\,\partial_5G(x)+o(s^2).
\]
For small $s>0$ the two points $x\pm sv-\delta s^2e_5$ therefore lie in $\{G>\epsilon\}$, while their midpoint
$x-\delta s^2e_5$ lies outside $\{G\ge\epsilon\}$, so neither set is convex. Because $\nabla G(x)^\top v=0$, also
$v^\top\nabla^2(\log G)(x)\,v=c/\epsilon>0$, so $\log G$ is not concave on any neighborhood of $x$.

Apply this with $G=\tilde F_9$ in (a) and $G=F_9$ in (b), and with $v=e_1-(\partial_1G/\partial_4G)(x)\,e_4$, so that
$\nabla G(x)^\top v=0$. Since $V_9=p_5C$ with $C>0$, $\partial_5G(x)$ equals $p_5(1-p_5)C>0$ in (a) and $(1-p_5)C>0$
in (b). The derivatives of $G$ follow from those of the polynomial $V_9$ by the chain rule, with
$dp_i/d\xi_i=p_i(1-p_i)$, $d^2p_i/d\xi_i^2=p_i(1-p_i)(1-2p_i)$, $dp_i/du_i=1-p_i$ and $d^2p_i/du_i^2=-(1-p_i)$, so at
a rational point they are rational numbers. Exact evaluation gives $\partial_1G/\partial_4G\approx0.99071$ and
$c/\epsilon\approx2.715\times10^{-10}$ in (a), and $\partial_1G/\partial_4G\approx0.99943$ and
$c/\epsilon^\circ\approx3.510\times10^{-14}$ in (b). Only the signs of $\partial_5G(x)$ and $c$ are used, and both are
exact. The supplement's \texttt{nonconvexity\_check.py} computes $V_9$ by the recursion, by enumerating all $2^9$
outcome sequences and by the formula for $C$, and checks each sign above in exact rational arithmetic; it is an exact
computation, not a proof-assistant verification.
\end{proof}

The two parts concern different points and thresholds. With $D_p=\mathrm{diag}(p)$, the chain rule gives
\[
  \nabla^2_\xi\log\tilde F_B=D_p\bigl(\nabla^2_u\log F_B\bigr)D_p+\mathrm{diag}\bigl(p_i(1-p_i)\,
  \partial\log F_B/\partial u_i\bigr),
\]
and $v\mapsto D_pv$ maps tangent vectors of $\{\tilde F_B=\epsilon\}$ to tangent vectors of $\{F_B=\epsilon\}$. The
diagonal term is nonnegative by Lemma~\ref{lem:domination}, so positive tangent curvature in $u$ gives positive tangent
curvature in $\xi$ at the same point, and part (b) also holds in logit coordinates. The converse fails: at the point
of (a), the Hessian of $F_9$ in $u$, restricted to the tangent space of $\{F_9=1/100\}$, is negative definite.

Both points have several nearly certain milestones, and the positive curvatures are of order $10^{-10}$ and $10^{-14}$.
The threshold $1/100$ in (a) results from the choice of $p_5$; it does not make the point a typical operating point.

The certificates of Theorems~\ref{thm:finite}, \ref{thm:cyclecert} and~\ref{thm:cyclecert-general} and of
Proposition~\ref{prop:crosscert} bound $V_B$ by functions whose logarithms are concave in $u$, so the superlevel sets on which
these bounds are at least $\epsilon$ are convex. Lemma~\ref{lem:zeropath}, Proposition~\ref{prop:sandwich} and the asymptotic
results of Section~\ref{sec:sharp} do not use convexity of $S$, and Proposition~\ref{prop:nonconvex} does not affect
them. It does not decide whether $\I^*=t$ in general (Example~\ref{ex:nonuniform} and Proposition~\ref{prop:schur}
give cases where equality holds). It also concerns only the sets and coordinates named in it, and says nothing about
other formulations of the problem or about computing $\sup_{u\in\Pi}F_B(u)$ in Section~\ref{sec:gdm}.

\section{Proofs for Section~\ref{sec:finite}}\label{app:finite}

\begin{proof}[Proof of Theorem~\ref{thm:finite}]
(i) The first inequality is Lemma~\ref{lem:envelope}(c). For the second, list the selected segments in chain order
$j_1<\dots<j_a$, let $\bar k_m$ be the number of milestones up to and including segment $j_m$, and on a successful
campaign let $b_m$ be the attempt at which segment $j_m$ is completed. Segment $j_m$ is completed by an uninterrupted
run of successes at its own milestones, occupying attempts $b_m-\ell_{j_m}+1,\dots,b_m$. Mandatory progress gives
$b_1\ge\bar k_1$, $b_{m+1}-b_m\ge\bar k_{m+1}-\bar k_m$ and $b_a\le B-(k-\bar k_a)$. Hence $x_m=b_m-\bar k_m$ is a
weakly increasing integer sequence in $[0,B-k]$, and there are exactly $\binom{B-k+a}{a}$ such sequences (stars and
bars). For a fixed sequence the success windows are disjoint, because $\bar k_{m+1}-\bar k_m\ge\ell_{j_{m+1}}$. Expose the history in time order: each
prescribed success has conditional probability at most the parameter of its milestone under (D3), or is impossible if
the process is at another state. Iterated conditioning bounds the probability of the pattern by $\prod_{E_A}p_i$, and
a union bound over sequences proves (i).

(ii) The function $g(u)=\log(1-e^{-u})$ satisfies $g''(u)=-e^u/(e^u-1)^2<0$ for $u>0$. For $u\in S$, (i) gives
$\epsilon/C_A(B)\le\prod_{E_A}(1-e^{-u_i})$, so every selected $u_i$ is positive and, by Jensen's inequality,
$\prod_{E_A}(1-e^{-u_i})\le(1-e^{-\bar u})^{K_A}$ with $\bar u=K_A^{-1}\sum_{E_A}u_i$. Hence $\bar u\ge\tau_A$, that is,
$w_A^\top u\ge\tau_A$. The other inequalities are Lemma~\ref{lem:closure}.

(iii) Under any $p$ with $V_B(p)>\epsilon$, (i) and (ii) give $\sum_{E_A}u_i\ge K_A\tau_A$ (if some selected
$p_i=1$, the likelihood below is zero), so $\Prob_p(\text{all zero})=e^{-n_A\sum_{E_A}u_i}\le e^{-n_AK_A\tau_A}\le
\alpha$. Because $A$ is fixed before the data, this bound is the probability of certification. At $p=0$ all responses
are zero, and the cost bound is $\lceil x\rceil<x+1$.
\end{proof}

\begin{proof}[Proof of Proposition~\ref{prop:plausibility}]
If a certified pair has $V_B(p)>\epsilon$, then $u(p)\notin\Pi$, that is, $\sum_in_iu_i(p)>c_\alpha$, and
certification requires all responses to be zero, which has probability $e^{-\sum_in_iu_i(p)}<\alpha$. The event does
not depend on the pair, so the bound holds simultaneously for all pairs.
\end{proof}

\section{Proofs and examples for Section~\ref{sec:sharp}}\label{app:sharp}

\begin{proof}[Proof of Lemma~\ref{lem:zeropath}]
Condition on the private randomization. Along the all-zero path the sequence of queried milestones and the almost
surely finite certification time are then fixed. Under a strictly unsafe finite $u$ the probability of following that
path is $e^{-\sum_iu_iN_i}$; certification may also occur on other paths, so validity gives
$\E_0e^{-\sum_iu_iN_i}\le\Prob_u(\text{certify})\le\alpha$. Jensen's inequality gives $\sum_iu_i\E_0N_i\ge c_\alpha$.
Extend to $S$ by continuity when the expected counts are finite, and take the infimum over $u\in S$. Since
$I_w\le\I^*$ the first claim follows; infinite expected cost is trivial.
\end{proof}

\begin{proof}[Proof of Proposition~\ref{prop:sandwich}]
For $u\in U$, $\sum_in_iu_i\ge(c_\alpha/\I^*)\,w^{*\top}u\ge c_\alpha$, so the all-zero probability is at most
$\alpha$, and summing $k$ ceilings costs less than $k$; the lower bound is Lemma~\ref{lem:zeropath}. For the second
statement, if $u\in S$ then $(\max_iu_i)\one\ge u$ lies in $S$ by Lemma~\ref{lem:envelope}(a), so $\max_iu_i\ge t$
and $k^{-1}\sum_iu_i\ge t/k$.
\end{proof}

\subsection{The cycle-length renewal lemma}\label{app:renewal}

Fix a segment $j$ with probabilities $p_{j,1},\dots,p_{j,\ell_j}$ in chain order, at positions $0,\dots,\ell_j-1$,
where a failure returns to position $0$. A cycle starts at position $0$ and ends at the first failure or at
completion. Its length $C$ satisfies $\Prob(C>h)=\pi_{j,h}$ for $0\le h<\ell_j$ and $C\le\ell_j$, so $\E C=\kappa_j$,
and it completes the segment with probability $\beta_j$. Let $T_j$ be the number of attempts the chain spends in
segment $j$ ($T_j=\infty$ if $\beta_j=0$). In the independent chain, each segment uses its own independent attempts, so $T_1,\dots,T_s$ are independent and
$V_B(p)=\Prob(\sum_jT_j\le B)$.

For counting purposes, define a hypothetical \emph{continued} process for segment $j$: a new cycle starts after
every cycle, including a completing one. The actual chain leaves the segment after completion. Write $U_j(m)$ for
the probability that a cycle starts at time $m$ in the continued process.

If $0<\beta_j<1$, define a second renewal process using independent failed-cycle lengths. Their distribution is that
of $C$ conditional on failure, with mean $\kappa^F_j=(\kappa_j-\ell_j\beta_j)/(1-\beta_j)$. Write $U^F_j(m)$ for
its probability of a cycle start at time $m$. We use the conventions $0\cdot\infty=0$ and $\kappa_j/\beta_j=\infty$
when $\beta_j=0$.

\begin{lemma}[Cycle lengths and the renewal sandwich]\label{lem:renewal-full}
\begin{enumerate}
\item[(a)] If $\beta_j>0$ then $\E T_j=\kappa_j/\beta_j$. For every integer $x\ge0$,
\[
  \frac{x+1}{\kappa_j}\le\sum_{m\le x}U_j(m)\le\frac{x+\ell_j}{\kappa_j},
  \qquad\text{and, if }0<\beta_j<1,\quad
  \sum_{m\le x}U^F_j(m)\ge\frac{x+1}{\kappa^F_j}\ge\frac{x+1}{\kappa_j}.
\]
\item[(b)] Let $B\ge k$, let $A$ be a nonempty set of segments with $a=|A|$ and $K_A=\sum_{j\in A}\ell_j$, and let
$Y=\sum_{j\notin A}T_j$. For every $p\in[0,1]^k$,
\[
  \prod_{j\in A}\frac{\beta_j(1-\beta_j)^B}{\kappa_j}\;
  \E\Bigl[\tbinom{B-Y-K_A+a}{a};\,Y\le B-K_A\Bigr]
  \;\le\; V_B(p)\;\le\;\frac{(B-k+K_A)^a}{a!}\prod_{j\in A}\frac{\beta_j}{\kappa_j}
  \;\le\;\frac{B^a}{a!}\prod_{j\in A}\frac{\beta_j}{\kappa_j}.
\]
\item[(c)] If $\beta_j>0$ for every $j\in A$, then
\[
  \Bigl(1-B\sum_{j\in A}\beta_j\Bigr)_+
  \Bigl(1-\frac aB\Bigl(K_A+\sum_{j\notin A}\frac{\kappa_j}{\beta_j}\Bigr)\Bigr)_+
  \le\frac{V_B(p)}{(B^a/a!)\prod_{j\in A}\beta_j/\kappa_j}\le1 .
\]
\item[(d)] Along any sequence with fixed segment lengths and fixed $A$ such that $B\to\infty$, $\beta_j>0$ for
$j\in A$, $B\max_{j\in A}\beta_j\to0$ and $\sum_{j\notin A}\kappa_j/(B\beta_j)\to0$,
$V_B(p)\sim(B^a/a!)\prod_{j\in A}\beta_j/\kappa_j$. No condition is placed on individual coordinates. If every
coordinate tends to zero, every $\kappa_j\to1$ and $V_B(p)\sim(B^a/a!)\prod_{j\in A}\beta_j$.
\end{enumerate}
Lemma~\ref{lem:renewal}(a) is the upper bound in (b), and Lemma~\ref{lem:renewal}(b) is (d).
\end{lemma}

\begin{proof}
(a) Cycles are independent and identically distributed, and the number of cycles up to and including the completing
one is geometric with mean $1/\beta_j$; Wald's identity gives $\E T_j=\kappa_j/\beta_j$. In the continued process let
$S_n$ be the end of the $n$-th cycle ($S_0=0$) and $N(x)=\#\{n\ge0:S_n\le x\}$ the number of cycle starts in
$\{0,\dots,x\}$, so that $\E N(x)=\sum_{m\le x}U_j(m)$. $N(x)$ is a stopping time for the cycle lengths, and
$x+1\le S_{N(x)}\le x+\ell_j$ because every cycle has length between $1$ and $\ell_j$; Wald's identity
$\E S_{N(x)}=\kappa_j\E N(x)$ gives the two-sided bound. Failed cycles also have length at least one, which gives the
bound for $U^F_j$, and $\kappa^F_j\le\kappa_j$ because $\kappa_j=\beta_j\ell_j+(1-\beta_j)\kappa^F_j$ and
$\kappa_j\le\ell_j$.

(b) \emph{Upper bound.} Segment $j$ completes at time $n$ only if a cycle of the continued process starts at
$n-\ell_j$ and completes; the completion of that cycle depends only on later attempts, so
$\Prob(T_j=n)\le\beta_jU_j(n-\ell_j)$. Since $T_j\ge\ell_j$ for $j\notin A$,
\[
  V_B(p)\le\Prob\Bigl(\sum_{j\in A}T_j\le B-k+K_A\Bigr)\le\prod_{j\in A}\beta_j\;G_A(B-k),\qquad
  G_A(x)=\sum_{m\in\mathbb N^A:\,\sum_jm_j\le x}\ \prod_{j\in A}U_j(m_j).
\]
We show $G_A(x)\le(x+K_A)^a/(a!\prod_A\kappa_j)$ by induction on $a$; the case $a=1$ is (a). For $a\ge2$ let $j^*$ be
the last segment of $A$, $A'=A\setminus\{j^*\}$, $f(y)=(y+K_{A'})^{a-1}/((a-1)!\prod_{A'}\kappa_j)$ the bound for $A'$
(nondecreasing, with $f(-1):=0$), and $R(y)=\sum_{m\le y}U_{j^*}(m)\le(y+\ell_{j^*})/\kappa_{j^*}$. Summation by parts
and the nonnegative increments of $f$ give
\[
  G_A(x)\le\sum_{y=0}^xU_{j^*}(y)f(x-y)=\sum_{y=0}^xR(y)\bigl[f(x-y)-f(x-y-1)\bigr]
  \le\frac1{\kappa_{j^*}}\Bigl[\ell_{j^*}f(x)+\sum_{y=0}^{x-1}f(y)\Bigr].
\]
Bounding the sum by $\int_0^xf$ and using $X^a+a\ell X^{a-1}\le(X+\ell)^a$ with $X=x+K_{A'}$ gives the claim, since
$K_{A'}+\ell_{j^*}=K_A$.

\emph{Lower bound.} If $\beta_j\in\{0,1\}$ for some $j\in A$, the left side is zero. Otherwise, completion at time $n$
is a run of $i$ failed cycles of total length $n-\ell_j$ followed by a completing cycle, and $i\le n$, so
$\Prob(T_j=n)\ge\beta_j(1-\beta_j)^nU^F_j(n-\ell_j)$. We again induct on $a$. The one-segment case follows from (a). For $a\ge2$, keep $j^*$ and $A'$ as above but use
\[
  R^F(y)=\sum_{m\le y}U^F_{j^*}(m)\ge\frac{y+1}{\kappa_{j^*}},\qquad
  f(y)=\frac{\binom{y+a-1}{a-1}}{\prod_{j\in A'}\kappa_j}\quad(y\ge0),\qquad f(-1)=0.
\]
The induction hypothesis and summation by parts give
\begin{align*}
  \sum_{\sum_jm_j\le x}\prod_{j\in A}U^F_j(m_j)
  &\ge\sum_{y=0}^xR^F(y)\bigl[f(x-y)-f(x-y-1)\bigr]\\
  &\ge\frac1{\kappa_{j^*}}\sum_{y=0}^xf(y)
   =\frac{\binom{x+a}{a}}{\prod_{j\in A}\kappa_j}.
\end{align*}
Here the hockey-stick identity replaces the integral used for the upper bound. Hence
$\Prob(\sum_AT_j\le M)\ge\prod_A\beta_j(1-\beta_j)^M\kappa_j^{-1}\binom{M-K_A+a}{a}$ for $M\ge K_A$, and conditioning
on the independent $Y$ with $M=B-Y$ proves the lower bound.

(c) Use $(1-\beta)^B\ge1-B\beta$, $\prod_j(1-x_j)\ge1-\sum_jx_j$ on $[0,1]$, $\binom{M+a}{a}\ge M^a/a!$ for $M\ge0$,
$(1-x)_+^a\ge1-ax$ for $x\ge0$, and $\E Y=\sum_{j\notin A}\kappa_j/\beta_j$ from (a).

(d) Both factors in (c) tend to one, since $B\sum_{j\in A}\beta_j\le aB\max_{j\in A}\beta_j\to0$, $K_A/B\to0$ and
$\sum_{j\notin A}\kappa_j/(B\beta_j)\to0$. If every coordinate tends to zero, then
$\kappa_j=1+\sum_{h\ge1}\pi_{j,h}\to1$.
\end{proof}

\subsection{Phase interiors}

\begin{proof}[Proof of Lemma~\ref{lem:active}]
If $j\in A$ and $\ell_{j'}\ge\ell_j$, then $B_np_n^{\ell_{j'}}\le B_np_n^{\ell_j}\to0$, whereas each $j'\notin A$ needs
$B_np_n^{\ell_{j'}}\to\infty$; so $j'\in A$. For the converse, $B_np_n^{\ell}\sim p_n^{\ell-\gamma}$, which tends to
zero for $\ell\ge\ell^*>\gamma$ and to infinity for $\ell\le\ell_-<\gamma$; there are finitely many free segments, and
$B_n\to\infty$ since $\gamma>0$. If all segments have the same length, the only active set is the whole chain; in the
last phase the segments of maximal length $L$ are active together.
\end{proof}

\begin{proof}[Proof of Theorem~\ref{thm:sharp}]
On the diagonal, $\beta_j=p_n^{\ell_j}$ and $\kappa_j=\sum_{h<\ell_j}p_n^h\to1$, so
$\sum_{j\notin A}\kappa_j/(B_n\beta_j)\le L\sum_{j\notin A}(B_np_n^{\ell_j})^{-1}\to0$, and
Lemma~\ref{lem:renewal-full}(d) gives $\epsilon_n\sim B_n^ap_n^{K_A}/a!$, which tends to zero because each active
$B_np_n^{\ell_j}$ does. As $C_A(B_n)\sim B_n^a/a!$, we get $r_{A,n}/p_n\to1$ and $\tau_{A,n}/t_n\to1$, and
Theorem~\ref{thm:finite}(ii) squeezes $\I^*_n$. By Lemma~\ref{lem:zeropath} and Theorem~\ref{thm:finite}(iii),
\[
  \frac{c_{\alpha_n}}{\I^*_n}\le\inf\E_0N<\frac{c_{\alpha_n}}{\tau_{A,n}}+K_A ,
\]
and the additive $K_A$ is negligible because $c_{\alpha_n}/t_n\to\infty$. If $c_{\alpha_n}/t_n$ stays bounded,
integrality can dominate and no constant is claimed.
\end{proof}

\begin{proof}[Proof of Theorem~\ref{thm:alloc}]
Fix $z\in(0,\infty)^k$. The probabilities $1-e^{-t_nz_i}\sim p_nz_i$ all tend to zero, so every $\kappa_j\to1$, and
they satisfy the hypotheses of Lemma~\ref{lem:renewal-full}(d) with the same $A$; hence
$F_{B_n}(t_nz)/\epsilon_n\to\prod_{i\in E_A}z_i$. If this product exceeds one, $t_nz$ is eventually strictly unsafe and
$I_{w_n,n}\le t_nw_n^\top z$. Along a subsequence with $w_n\to w$, the hypothesis gives $w^\top z\ge1$ for every
positive $z$ whose active product exceeds one. First take $n\to\infty$ with $z$ fixed, as above. Then let the active product decrease to one and the free
coordinates decrease to zero. This order avoids requiring uniform convergence in $z$ and gives
\[
  1\le\inf_{\substack{z_i>0\,(i\in E_A)\\ \prod_{E_A}z_i=1}}\ \sum_{i\in E_A}w_iz_i .
\]
If some active weight is zero, this infimum is zero (send that coordinate to infinity and the others to zero; if
$K_A=1$ the sum is zero outright), a contradiction. Otherwise the infimum equals $K_A(\prod_{E_A}w_i)^{1/K_A}$ by the
weighted inequality of arithmetic and geometric means, and the same inequality gives
$K_A(\prod_{E_A}w_i)^{1/K_A}\le\sum_{E_A}w_i\le1$. Equality throughout forces $w_i=1/K_A$ on $E_A$ and $w_i=0$
elsewhere; all subsequential limits coincide, which proves (a). For (b), Lemma~\ref{lem:zeropath} gives
$I_{w_n,n}\ge c_{\alpha_n}/\E_0N\sim t_n$, while $I_{w_n,n}\le\I^*_n\le t_n$.
\end{proof}

\subsection{Finite budgets, full retention and boundary cases}

\begin{example}[Finite optimal allocation need not be uniform]\label{ex:nonuniform}
Take one segment of length two and $B=3$. Completion within three attempts requires either no failure or one failure
at the first milestone, so $F_3(u)=(1-e^{-2u_1})(1-e^{-u_2})$. $\log F_3$ is a sum of strictly concave functions of
$u_1$ and $u_2$, so $S$ is convex. At $t\one$, concavity gives $\nabla\log F_3(t\one)^\top(u-t\one)\ge0$ for all
$u\in S$, so the normalized gradient $w^*$ satisfies $I_{w^*}\ge t$, hence $\I^*=t$ by Lemma~\ref{lem:closure}, with
$w^*_1/w^*_2=2q/(1+q)<1$ for $q=e^{-t}$. The normal of the boundary at $t\one$ is not parallel to $\one$, so a tangent direction decreases $\one^\top u$.
Since $\partial_{u_2}F_3>0$, the implicit function theorem gives a $C^1$ boundary curve in that direction.
A nearby boundary point therefore has $\one^\top u/2<t$, making uniform weights strictly suboptimal at this budget. Numerically, at $p_*=0.1$ the optimal weight is $w_1^*=0.4865$ and
$I_{\one/2}=0.105325<\I^*=0.105361$. Theorem~\ref{thm:alloc} says only that the optimal proportions tend to
$(1/2,1/2)$ along phase-interior sequences; the numerical optimum is $w_1^*=0.4929$ at $(B,p_*)=(30,0.03)$ and $0.4993$
at $(300,0.003)$.
\end{example}

\begin{proposition}[Full progress retention: uniform allocation is exact at every budget]\label{prop:schur}
If $\ell_j=1$ for every $j$, then for every $B\ge k$ and $0<\epsilon<1$, $F_B$ is Schur-concave in $u$,
$I_{\one/k}=\I^*=t$, and $c_\alpha/t\le\inf\E_0N<c_\alpha/t+k$.
\end{proposition}

\begin{proof}
Here $V_B(p)=\Prob(\sum_iG_i\le B)$ for independent geometric variables $G_i$ on $\{1,2,\dots\}$ with
$\Prob(G_i>n)=q_i^n$, $q_i=e^{-u_i}$. For two variables with $q_1=ae^\delta$ and $q_2=ae^{-\delta}$ (so that
$u_1+u_2$ is fixed),
\[
  \Prob(G_1+G_2>n)=\frac{(1-q_2)q_1^n-(1-q_1)q_2^n}{q_1-q_2}
  =a^{n-1}\bigl[H_n(\delta)-aH_{n-1}(\delta)\bigr],\qquad H_n(\delta)=\frac{\sinh n\delta}{\sinh\delta}.
\]
Writing $H_n=\sum_{j=0}^{n-1}e^{(n-1-2j)\delta}$, we have $H_n'=2\sum m\sinh(m\delta)$ over $0<m\le n-1$ with
$m\equiv n-1\pmod2$. Matching each term $m$ of $H_{n-1}'$ with the term $m+1$ of $H_n'$ (the extra term $m=1$ when
$n$ is even is positive) and using that $m\sinh(m\delta)$ increases in $m$ gives $H_n'>H_{n-1}'\ge0$ for $n\ge2$ and
$\delta>0$; since $a<1$, the survival function increases with $\delta$. Thus spreading $(u_1,u_2)$ at a fixed sum
lowers the two-variable distribution function. For $k$ variables, condition on $R=\sum_{i\ge3}G_i$: the distribution
function is a mixture, with weights free of $(u_1,u_2)$, of two-variable distribution functions, so every pairwise
equalization raises it; with symmetry, this is Schur-concavity. For $u\in S$, $F_B(\bar u\one)\ge F_B(u)\ge\epsilon$,
so $\bar u\ge t$ and $I_{\one/k}\ge t$; Lemma~\ref{lem:closure} and Proposition~\ref{prop:sandwich} finish the proof.
\end{proof}

\begin{example}[Cycle costs without vanishing rates]\label{ex:cyclecost}
(a) Take one segment of length two with $p_1=1/4$ fixed and $p_2=B^{-2}$. Then $B\beta\to0$, but
$V_B/(Bp_1p_2)\to4/5$. This is Lemma~\ref{lem:renewal-full}(d) with $\kappa=1+p_1=5/4$: the cycle-free expression
$(B^a/a!)\prod_A\beta_j$ is correct only when $\kappa_j\to1$ on the active segments, for instance when all rates tend
to zero, whereas with the cycle means no condition on individual rates is needed. (b) With $p_1=1$ (a sure first step), every cycle has length exactly two, so $T=2G$ with $G$ geometric with
parameter $q=p_2$, and $V_B=1-(1-q)^{\lfloor B/2\rfloor}$ exactly. As $B\to\infty$ with $Bq\to0$, $V_B\sim Bq/2$, half the window
coefficient $B-1$ of Theorem~\ref{thm:finite}. In the unrestricted class such points do not determine the zero-world
constant (Remark~\ref{rem:scope}). They do in classes in which $p_1$ is fixed or capped: with $Q=\{2\}$ the problem is
one-dimensional, the restricted diagonal threshold is $t=-\log(1-\epsilon)/\lfloor B/2\rfloor$, and the window
certificate costs about twice the optimum.
\end{example}

\begin{example}[Fixed risk, one milestone]\label{ex:fixedeps}
For $k=1$, $V_B(p)=1-(1-p)^B$, so $t=-\log(1-\epsilon)/B$, and Proposition~\ref{prop:sandwich} gives
$\inf\E_0N\in[c_\alpha B/(-\log(1-\epsilon)),\,c_\alpha B/(-\log(1-\epsilon))+1)$, not $c_\alpha B/\epsilon$. The
$\epsilon$ of Theorem~\ref{thm:sharp} is a rare-risk asymptotic quantity.
\end{example}

\begin{corollary}[Full reset and full retention]\label{cor:corners}
Fix $\alpha$, let $B_n\to\infty$ and $p_n\to0$, and put $\epsilon_n=V_{B_n}(p_n\one)$. Under full reset with
$B_np_n^k\to0$, $\inf\E_0N\sim c_\alpha(B_n/\epsilon_n)^{1/k}$; under full progress retention with $B_np_n\to0$,
$\inf\E_0N\sim c_\alpha B_n(k!\,\epsilon_n)^{-1/k}$. In both cases the limiting allocation is uniform.
\end{corollary}

\begin{proof}[Proof of Corollary~\ref{cor:corners}]
Under full reset the only set of segments is the whole chain, with $a=1$ and $K_A=k$, and $B_np_n^k\to0$ makes
$(B_n,p_n)$ a diagonal phase-interior sequence for it. Under full retention every segment has length one, so by
Lemma~\ref{lem:active} the only active set is the set of all segments, with $a=K_A=k$, and the condition is
$B_np_n\to0$. Theorems~\ref{thm:sharp} and~\ref{thm:alloc} give the constants and the uniform limiting allocation.
\end{proof}

\subsection{A single crossover}\label{app:crossover}

Throughout this subsection $M$ is one segment, of length $m$ and with milestones $E_M$, $A$ is a nonempty set of
segments not containing $M$, $a=|A|$, and the remaining segments are free in the asymptotic statements. For
$\lambda\in(0,\infty]$ let
\[
  h_a(\lambda)=\E\bigl(1-E/\lambda\bigr)_+^a=\E\bigl[1-e^{-\lambda V}\bigr],\qquad
  e_a(\lambda)=\frac{\lambda h_a'(\lambda)}{h_a(\lambda)},
\]
where $E\sim\mathrm{Exp}(1)$ and $V\sim\mathrm{Beta}(1,a)$ is independent of $E$, so that $\Prob(V>x)=(1-x)_+^a$; put
$h_a(\infty)=1$ and $h_a(0)=0$. For $a=1$, $h_1(\lambda)=1-(1-e^{-\lambda})/\lambda$ and
$e_1(\lambda)=(1-(1+\lambda)e^{-\lambda})/(\lambda-1+e^{-\lambda})$. As $\lambda\downarrow0$,
$h_a(\lambda)\sim\lambda\E V=\lambda/(a+1)$.

On the diagonal, part (a) of Theorem~\ref{thm:crossover-full} below describes the risk function only. It follows
from the exponential limit of geometric sums \citep{renyi1956characterization}, and we do not claim it as new; the
proof below is included for completeness. The evidence-complexity results need, in addition, the
finite bound of Lemma~\ref{lem:crossbound}, the log-concavity of Lemma~\ref{lem:profile}, and the limit at the
off-diagonal points $t_nz$.

\begin{lemma}[An exponential inequality]\label{lem:expineq}
For $x,y\in[0,\infty]$, with $0\cdot\infty=0$, $(1-e^{-x})(1-e^{-y})\le1-e^{-xy}$. Consequently
$-\log(1-\beta_j)\le\prod_{i\in E_j}u_i$ for every segment $j$.
\end{lemma}

\begin{proof}
If $x$ or $y$ is $0$ or $\infty$ the claim is immediate. Let $0<x,y<\infty$. If $x\ge1$, then
$1-e^{-xy}\ge1-e^{-y}\ge(1-e^{-x})(1-e^{-y})$. If $x<1$, the function $s\mapsto1-e^{-s}$ is concave and vanishes at
$0$, so $1-e^{-xy}\ge x(1-e^{-y})\ge(1-e^{-x})(1-e^{-y})$. Induction on the length of the segment gives
$\beta_j=\prod_{E_j}(1-e^{-u_i})\le1-e^{-\prod_{E_j}u_i}$.
\end{proof}

\begin{lemma}[Finite crossover bound]\label{lem:crossbound}
Let $B\ge k$ and $B_\times=B-k+a+1$. For every $p\in[0,1]^k$ and every process in $\Dcl(p)$,
\[
  \Prob(\text{completion within }B)\le V_B(p)\le\prod_{i\in E_A}p_i\cdot\frac{B_\times^a}{a!}\,
  h_a\Bigl(B_\times\prod_{i\in E_M}u_i\Bigr).
\]
\end{lemma}

\begin{proof}
The first inequality is Lemma~\ref{lem:envelope}(c). Let $J$ be the segments outside $A\cup\{M\}$ and $K_J$ their
number of milestones. Since $T_j\ge\ell_j$, $V_B(p)\le\Prob(\sum_AT_j\le B-K_J-T_M)$. Theorem~\ref{thm:finite}(i),
applied to the chain formed by the segments of $A$ alone with all of them selected, gives
$\Prob(\sum_AT_j\le y)\le c(y-K_A+a)\prod_{E_A}p_i$ for every integer $y$, where $c(n)=\binom na$ for integers $n\ge a$
and $c(n)=0$ for $n<a$; $c$ is nondecreasing. By independence,
$V_B(p)\le\prod_{E_A}p_i\;\E\,c(B-K_J-T_M-K_A+a)$. If $\beta_M=0$ then $T_M=\infty$ and both sides vanish. The case $\beta_M=1$ follows by continuity, so assume $0<\beta_M<1$. Then
$T_M$ is $m$ plus the lengths of the $G-1$ failed cycles, where $G\sim\mathrm{Geom}(\beta_M)$ on $\{1,2,\dots\}$, and
each failed cycle has length at least one, so $T_M\ge G-1+m$ and the argument of $c$ is at most $B_\times-G$. Couple
$G=\lceil E/\eta\rceil\ge E/\eta$ with $\eta=-\log(1-\beta_M)$. For integers $n\le x$, $c(n)\le x_+^a/a!$ because
$\binom na\le n^a/a!$, so
\[
  \E\,c(B_\times-G)\le\frac{1}{a!}\,\E\Bigl(B_\times-\frac E\eta\Bigr)_+^a=\frac{B_\times^a}{a!}\,h_a(B_\times\eta).
\]
Since $h_a$ is nondecreasing, Lemma~\ref{lem:expineq} finishes the proof.
\end{proof}

\begin{lemma}[The crossover profile]\label{lem:profile}
The function $s\mapsto\log h_a(e^s)$ is concave and nondecreasing on $\mathbb R$. Consequently $e_a$ is nonincreasing
on $(0,\infty)$, takes values in $(0,1)$, and satisfies $e_a(0+)=1$ and $e_a(\infty)=0$; and for every $B_\times>0$,
$u\mapsto\log h_a(B_\times\prod_{i\in E_M}u_i)$ is concave on $(0,\infty)^{E_M}$.
\end{lemma}

\begin{proof}
Let $\lambda=e^s$ and $W=\log E-s$, which has density $g(w+s)$ with $g(x)=\exp(x-e^x)$; then $E/\lambda=e^W$ and
\[
  h_a(e^s)=\Prob(e^W<V)=\int\one\{e^w<v<1\}\,g(w+s)\,a(1-v)^{a-1}\,dw\,dv .
\]
The integrand is log-concave in $(s,w,v)$: the set $\{e^w<v<1\}$ is convex, $\log g(w+s)=w+s-e^{w+s}$ is concave, and
$(1-v)^{a-1}$ is log-concave on $(0,1)$ for $a\ge1$. By Pr\'ekopa's theorem \citep{prekopa1973logarithmic}, the marginal
in $s$ is log-concave; it is nondecreasing because $E/\lambda$ decreases in $\lambda$. Hence
$e_a(\lambda)=\frac{d}{ds}\log h_a(e^s)$ is nonincreasing. Since $0<xe^{-x}<1-e^{-x}$ for $x>0$,
$e_a(\lambda)=\E[\lambda Ve^{-\lambda V}]/\E[1-e^{-\lambda V}]\in(0,1)$; as $\lambda\downarrow0$ numerator and
denominator are both $\sim\lambda\E V$, and as $\lambda\to\infty$ the numerator tends to $0$ by dominated convergence
while the denominator tends to $1$. Finally, $\log h_a(B_\times\prod_{E_M}u_i)=H(\log B_\times+\sum_{E_M}\log u_i)$ with
$H(s)=\log h_a(e^s)$ concave and nondecreasing and $\sum_{E_M}\log u_i$ concave.
\end{proof}

\begin{proof}[Proof of Proposition~\ref{prop:crosscert}]
(i) is Lemma~\ref{lem:crossbound}. For (ii), $D(\tau)=(1-e^{-\tau})^{K_A}(B_\times^a/a!)\,h_a(B_\times\tau^m)$ is continuous and
strictly increasing from $0$ to $B_\times^a/a!\ge1>\epsilon$ (note $B_\times\ge a+1$), so $\tau_\times$ exists and is unique. Let
$F_B(u)\ge\epsilon$. By (i), $\epsilon\le\Phi(u):=\prod_{E_A}(1-e^{-u_i})\,(B_\times^a/a!)\,h_a(B_\times\prod_{E_M}u_i)$. If some
$u_i=0$ with $i\in E_A\cup E_M$, then $\Phi(u)=0$, which is impossible; if some such $u_i=\infty$, then $w^\top u=\infty$.
Otherwise $f=\log\Phi$ is concave and differentiable on $(0,\infty)^{E_A\cup E_M}$, by the concavity of
$\log(1-e^{-u})$ and Lemma~\ref{lem:profile}. At $u=\tau_\times\one$, $f=\log\epsilon$ and $\nabla f$ equals
$1/(e^{\tau_\times}-1)$ on $E_A$ and $e_a(B_\times\tau_\times^m)/\tau_\times$ on $E_M$, which is proportional to $w$. Concavity
gives $\log\epsilon\le f(u)\le\log\epsilon+\nabla f^\top(u-\tau_\times\one)$, hence $w^\top u\ge\tau_\times$, strictly
when $F_B(u)>\epsilon$. Thus $\tau_\times\le I_w$, and $I_w\le\I^*\le t$ by Lemma~\ref{lem:closure}. For validity, let
$V_B(p)>\epsilon$. If some queried $p_i=1$ the all-zero probability is zero; otherwise it is
$e^{-\sum n_iu_i}\le e^{-(c_\alpha/\tau_\times)w^\top u}<\alpha$. The design is fixed before any response, so this is
the certification probability. At $p=0$ every response is zero, and $\lceil x\rceil<x+1$ gives the cost bound.
\end{proof}

In the asymptotic statements $M$ is the only segment of its length $m$, $A=\{j:\ell_j>m\}\neq\emptyset$, and
$(B_n,p_n)$ is a \emph{single-crossover sequence}: integers $B_n\ge k$ with $B_n\to\infty$, $p_n\in(0,1)$ with
$p_n\to0$, and $B_np_n^m\to\lambda\in(0,\infty)$. Then $B_np_n^\ell\to0$ for $\ell>m$ and $B_np_n^\ell\to\infty$ for
$\ell<m$. Put $\epsilon_n=V_{B_n}(p_n\one)$ and $t_n=-\log(1-p_n)$, index $I_w$, $\I^*$ and $\tau_\times$ (with
$(A,M)$) by $(B_n,\epsilon_n)$, and let $\omega(\lambda)\in\Delta_k$ be the limit allocation of
Theorem~\ref{thm:crossover}.

\begin{theorem}[Single crossover, full statement]\label{thm:crossover-full}
Let $(B_n,p_n)$ be a single-crossover sequence.
\begin{enumerate}
\item[(a)] For every $z\in(0,\infty)^k$,
$F_{B_n}(t_nz)\sim(B_n^ap_n^{K_A}/a!)\prod_{E_A}z_i\,h_a(\lambda\prod_{E_M}z_i)$. In particular
$\epsilon_n\sim(B_n^ap_n^{K_A}/a!)\,h_a(\lambda)\to0$ and
$F_{B_n}(t_nz)/\epsilon_n\to G_\lambda(z):=\prod_{E_A}z_i\,h_a(\lambda\prod_{E_M}z_i)/h_a(\lambda)$.
\item[(b)] $\tau_{\times,n}\sim\I^*_n\sim t_n\sim\bigl(a!\,\epsilon_n/(B_n^ah_a(\lambda))\bigr)^{1/K_A}\sim(\lambda/B_n)^{1/m}$.
\item[(c)] If $c_{\alpha_n}/t_n\to\infty$, then
$\inf_{\text{valid at }(B_n,\epsilon_n,\alpha_n)}\E_0N\sim c_{\alpha_n}\bigl(B_n^ah_a(\lambda)/(a!\,\epsilon_n)\bigr)^{1/K_A}$,
and the certificate of Proposition~\ref{prop:crosscert} with $(A,M)$ attains this leading constant.
\item[(d)] Every sequence $w_n\in\Delta_k$ with $I_{w_n,n}/t_n\to1$ satisfies $w_n\to\omega(\lambda)$. Hence the
allocation weights $\E_0N_i/\E_0N$ of every sequence of valid evaluators with $\E_0N\sim c_{\alpha_n}/t_n$ tend to $\omega(\lambda)$,
and so do the weights of Proposition~\ref{prop:crosscert}.
\end{enumerate}
Theorem~\ref{thm:crossover} consists of (a) at $z=\one$, $\I^*_n\sim t_n$, (c), and the evaluator part of (d).
\end{theorem}

\begin{proof}
(a) Fix $z$ and let $q_i=1-e^{-t_nz_i}\sim p_nz_i$. Every coordinate tends to zero, so every $\kappa_j\to1$. For
$j\in A$, $B_n\beta_j\to0$; for $M$, $B_n\beta_M\to\gamma:=\lambda\prod_{E_M}z_i\in(0,\infty)$; for every other
segment $j$, $B_n\beta_j\to\infty$. \emph{Upper half.} Lemma~\ref{lem:crossbound} gives
$F_{B_n}(t_nz)\le\prod_{E_A}q_i\,(B_{\times,n}^a/a!)\,h_a(B_{\times,n}t_n^m\prod_{E_M}z_i)$ with $B_{\times,n}\sim B_n$ and
$B_{\times,n}t_n^m\prod_{E_M}z_i\to\gamma$, and $h_a$ is continuous. \emph{Lower half.} Lemma~\ref{lem:renewal-full}(b) with
selected set $A$ and $Y=T_M+\sum_{j\notin A\cup\{M\}}T_j$ gives
\[
  F_{B_n}(t_nz)\ge\prod_{j\in A}\frac{\beta_j(1-\beta_j)^{B_n}}{\kappa_j}\;
  \E\Bigl[\tbinom{B_n-Y-K_A+a}{a};\,Y\le B_n-K_A\Bigr],
\]
where $(1-\beta_j)^{B_n}\ge1-B_n\beta_j\to1$ and $\kappa_j\to1$. For integers $y\ge0$,
$\binom{B-y-K_A+a}{a}\one\{y\le B-K_A\}/(B^a/a!)-(1-y/B)_+^a\to0$ uniformly in $y$: for $y\le B-K_A$ each of the $a$
factors $(B-y-K_A+i)/B$ differs from $1-y/B$ by $O(1/B)$, and all are at most $2$; for $y>B-K_A$,
$(1-y/B)_+^a\le(K_A/B)^a$. It therefore suffices that $Y/B_n\Rightarrow E/\gamma$, since then
$\E(1-Y/B_n)_+^a\to\E(1-E/\gamma)_+^a=h_a(\gamma)$, the function $x\mapsto(1-x)_+^a$ being bounded and continuous.
Let $G$ be the number of cycles of $M$, geometric with parameter $\beta_M$, so that $T_M\ge G$ and, by
Lemma~\ref{lem:renewal-full}(a), $\E(T_M-G)/B_n=(\kappa_M-1)/(B_n\beta_M)\to0$. Since
$\Prob(\beta_MG>x)=(1-\beta_M)^{\lfloor x/\beta_M\rfloor}\to e^{-x}$, $G/B_n=\beta_MG/(B_n\beta_M)\Rightarrow E/\gamma$;
and $\E T_j/B_n=\kappa_j/(B_n\beta_j)\to0$ for the free segments. Hence $Y/B_n\Rightarrow E/\gamma$. Finally
$\epsilon_n\to0$ because $B_n^ap_n^{K_A}=\prod_AB_np_n^{\ell_j}\to0$.

(b) Write $D_n$ for the function $D$ of Proposition~\ref{prop:crosscert} at $B_n$. For fixed $\rho>0$, (a) gives
$D_n(\rho p_n)/\epsilon_n\to\rho^{K_A}h_a(\lambda\rho^m)/h_a(\lambda)$, which is strictly increasing in $\rho$ and equals
$1$ at $\rho=1$. Since $D_n$ is increasing and $D_n(\tau_{\times,n})=\epsilon_n$, eventually $\tau_{\times,n}>\rho p_n$
for each $\rho<1$ and $\tau_{\times,n}<\rho p_n$ for each $\rho>1$, so $\tau_{\times,n}\sim p_n\sim t_n$. Proposition
\ref{prop:crosscert}(ii) gives $\tau_\times\le\I^*\le t$. The other forms follow from (a) and $B_np_n^m\to\lambda$.

(c) Lemma~\ref{lem:zeropath} and Proposition~\ref{prop:crosscert}(ii) give
$c_{\alpha_n}/\I^*_n\le\inf\E_0N<c_{\alpha_n}/\tau_{\times,n}+K_A+m$; apply (b), and note that $K_A+m$ is negligible
because $c_{\alpha_n}/t_n\to\infty$.

(d) We first show that all limiting weight lies on $E_A\cup E_M$, then identify its proportions from the gradient.
Let $w$ be a subsequential limit of $w_n$. If $G_\lambda(z)>1$, then by (a) $t_nz$ is eventually strictly unsafe,
so $I_{w_n,n}\le t_nw_n^\top z$ and, in the limit, $w^\top z\ge1$. $G_\lambda$ does not depend on the coordinates
outside $E_A\cup E_M$; letting them decrease to zero after $n\to\infty$ gives $\sum_{E_A\cup E_M}w_iz_i\ge1$ whenever
$G_\lambda(z)>1$. Let $g=\nabla\log G_\lambda(\one)$ on $E_A\cup E_M$, so $g_i=1$ on $E_A$ and $g_i=e_a(\lambda)$ on
$E_M$. If $g^\top v>0$, then $G_\lambda(\one+sv)>1$ for small $s>0$, so $W+s\,w^\top v\ge1$ with
$W=\sum_{E_A\cup E_M}w_i\le1$. Letting $s\downarrow0$ gives $W=1$, so $w$ vanishes off $E_A\cup E_M$, and then
$w^\top v\ge0$ whenever $g^\top v>0$.

We next show proportionality to the gradient. The last inequality forces $w=\mu g$ with $\mu\ge0$: otherwise write $w=\mu g+w_\perp$ with
$w_\perp\perp g$, $w_\perp\ne0$, and take $v=-w_\perp+\delta g$ with $\delta>0$ small.

Finally, the normalization $W=1$ gives
$w=\omega(\lambda)$, and all subsequential limits coincide. For valid evaluators, Lemma~\ref{lem:zeropath} gives
$I_{w_n,n}\ge c_{\alpha_n}/\E_0N\sim t_n$, while $I_{w_n,n}\le t_n$. For the certificate's weights,
$\tau_\times/(e^{\tau_\times}-1)\to1$ and $e_a(B_{\times,n}\tau_\times^m)\to e_a(\lambda)$.
\end{proof}

\begin{corollary}[The window certificate at a crossover]\label{cor:crosswindow}
Along a single-crossover sequence, let $A'$ be any set of segments chosen as a function of $(B_n,\epsilon_n)$. Then
\[
  \limsup_n\frac{\tau_{A',n}}{t_n}\le\max\Bigl\{h_a(\lambda)^{1/K_A},\ \Bigl(\frac{(a+1)\,h_a(\lambda)}{\lambda}\Bigr)^{1/(K_A+m)}\Bigr\}<1,
\]
so every certificate of Theorem~\ref{thm:finite} costs at least a constant factor more than the optimum. The two terms are the
limits for $A'=A$ and $A'=A\cup\{M\}$, and every other choice gives $\tau_{A'}/t_n\to0$. Every fixed allocation $w$
other than $\omega(\lambda)$, including the uniform allocations on $E_A$ and on $E_A\cup E_M$, has
$\limsup_nI_{w,n}/t_n<1$.
\end{corollary}

\begin{proof}
By Theorem~\ref{thm:crossover-full}(a), $C_{A'}(B_n)\sim B_n^{a'}/a'!$, and
\[
  \Bigl(\frac{r_{A',n}}{p_n}\Bigr)^{K_{A'}}\sim\frac{a'!}{a!}\,h_a(\lambda)
  \prod_{j\in A\setminus A'}B_np_n^{\ell_j}\Big/\prod_{j\in A'\setminus A}B_np_n^{\ell_j},
\]
which tends to zero unless $A\subseteq A'\subseteq A\cup\{M\}$; the two remaining choices give the displayed limits,
and $\tau_{A'}\sim r_{A'}$. Both are below one, because $h_a<1$ and
$h_a(\lambda)=\E[1-e^{-\lambda V}]<\lambda\E V=\lambda/(a+1)$. If a subsequence had $I_{w,n}/t_n\to1$,
Theorem~\ref{thm:crossover-full}(d) would force $w=\omega(\lambda)$.
\end{proof}

For segments $(3,1)$ the bound of Corollary~\ref{cor:crosswindow} is smallest, $0.851$, at $\lambda\approx2.35$, where
its two terms are equal, so the best window certificate then needs about $1.18$ times the optimal number of responses; over the geometries $(2,1)$,
$(3,1)$, $(3,2)$, $(5,1)$ and $(2,2,1)$ this worst-case factor ranges from $1.11$ to $1.26$.

\begin{remark}[Finite budgets and scope]\label{rem:crossscope}
(a) The certificate of Proposition~\ref{prop:crosscert} is not uniformly better than those of
Theorem~\ref{thm:finite}. Near $B=k$ its coefficient exceeds the window counts: for segments $(3,1)$ at
$\epsilon=10^{-4}$ and $\alpha=1/40$ it needs $44$ responses at $B=4$, against $36$ for the best window certificate.
All these certificates are computed from $(B,\epsilon,\alpha)$ before any response, so using the smallest of them is a
prospective choice and remains valid; Figure~\ref{fig:complexity} displays this minimum. (b) Theorem~\ref{thm:crossover}
is asymptotic. It identifies the leading constant and the limit of the optimal allocation weights along
single-crossover sequences, not the optimal design or its cost at a finite budget; Proposition~\ref{prop:crosscert} is
a valid certificate at every budget, not a finite-budget optimum. (c) The asymptotic optimality result does not cover two or more segments of the crossover length, which cross over
together on the diagonal; sequences along which $B_np_n^m$ does not converge to a positive finite limit;
heterogeneous milestone probabilities; restricted classes; or fixed $\epsilon$. These exclusions do not restrict
the validity of the finite certificate.
\end{remark}

\section{Cycle-aware certification and restricted classes: general statements and proofs}\label{app:restricted}

\subsection{Parameter classes}\label{app:classes}

A restricted class is specified by a nonempty set $Q\subseteq\{1,\dots,k\}$ of \emph{queryable} milestones, at which
Assumption~\ref{ass:O} holds, and a declared constant $c_i\in(0,1]$ for each $i\notin Q$. Such a coordinate is either
\emph{fixed} ($p_i=c_i$ is known) or \emph{capped} ($p_i\in[0,c_i]$ is unknown, and unrestricted if $c_i=1$).
Queryable coordinates range over $[0,1]$. Validity over a class takes the supremum of Section~\ref{sec:model} over the
unsafe parameters of the class only, and the zero world is $p_Q=0$. Put $\tilde c_i=c_i$ for $i\notin Q$ and
$\tilde c_i=1$ for $i\in Q$.

A class $\mathcal P_{Q,c}$ consists of the parameters $p\in[0,1]^k$ with $p_Q$ arbitrary and, for each $i\notin Q$,
$p_i=c_i$ if coordinate $i$ is fixed and $p_i\in[0,c_i]$ if it is capped. Write $\mathcal P^{\rm F}_{Q,c}$ when every
unqueryable coordinate is fixed and $\mathcal P^{\rm C}_{Q,c}$ when every one is capped, so that $\mathcal
P^{\rm F}_{Q,c}\subseteq\mathcal P_{Q,c}\subseteq\mathcal P^{\rm C}_{Q,c}$; the unrestricted class is $Q=\{1,\dots,k\}$.
Every $p\in\mathcal P_{Q,c}$ satisfies $p\le\tilde c$ coordinatewise. A zero declared constant would make $V_B\equiv0$
on the class and is excluded. A design may leave a queryable milestone unqueried; in a bound, such a milestone is
treated as capped at one.

At fixed queryable probabilities, setting all unqueryable coordinates to their declared caps maximizes risk by
monotonicity, without changing the audit observations. This is why the fixed, capped and mixed classes below yield
the same certification problem.

For $u\in[0,\infty)^Q$ let $(p(u),c)$ be the parameter with $p_i=1-e^{-u_i}$ for $i\in Q$ and $p_i=c_i$ for $i\notin
Q$, and put $F^Q_B(u)=V_B(p(u),c)$ and $\bar\epsilon_Q=V_B(\one_Q,c)$. By Lemma~\ref{lem:envelope}(a),
$F^Q_B(u)=\sup\{V_B(p):p\in\mathcal P_{Q,c},\ p_Q=p(u)\}$, the supremum being attained at the declared constants.
Because the responses depend on $p_Q$ alone, an evaluator is valid over $\mathcal P_{Q,c}$ if and only if it certifies
with probability at most $\alpha$ under every $p_Q\in[0,1]^Q$ with $V_B(p_Q,c)>\epsilon$. Validity, the zero world and
$\E_0N$ therefore depend on the class only through $(Q,c)$, and they are the same for $\mathcal P^{\rm F}_{Q,c}$,
$\mathcal P^{\rm C}_{Q,c}$ and every class between them. Define
\[
  U^Q=\{u:F^Q_B(u)>\epsilon\},\quad S^Q=\{u:F^Q_B(u)\ge\epsilon\},\quad
  I^Q_w=\inf_{u\in S^Q}w^\top u\ (w\in\Delta_Q),\quad \I^Q_*=\max_{w\in\Delta_Q}I^Q_w .
\]
For $Q=\{1,\dots,k\}$ these are the objects of Section~\ref{sec:model}. If $\epsilon\ge\sup_{p\in\mathcal P}V_B(p)$, the
class contains no unsafe parameter and the rule that certifies before any response is valid; information quantities
are then not used.

\begin{lemma}[Restricted closure and diagonal]\label{lem:rclosure}
Let $B\ge k$.
\begin{enumerate}
\item[(a)] $\lambda\mapsto V_B(\lambda\one_Q,c)$ is strictly increasing on $[0,1]$, from $0$ to $\bar\epsilon_Q>0$,
and $\bar\epsilon_Q=\sup_{p\in\mathcal P_{Q,c}}V_B(p)$.
\item[(b)] If $\epsilon\ge\bar\epsilon_Q$, the class contains no unsafe parameter, and certifying before any response
is valid.
\item[(c)] If $0<\epsilon<\bar\epsilon_Q$, then $S^Q=\overline{U^Q}$; there is a unique $t_Q>0$ with
$F^Q_B(t_Q\one_Q)=\epsilon$; $\I^Q_*$ is attained; and $\I^Q_*\le t_Q$.
\end{enumerate}
\end{lemma}

\begin{proof}
(a) $V_B(\lambda\one_Q,c)$ is a polynomial in $\lambda$, nondecreasing by Lemma~\ref{lem:envelope}(a). It vanishes at
$\lambda=0$ because $Q\neq\emptyset$ and a milestone with success probability zero cannot be passed. At $\lambda=1$
the path of $k$ consecutive successes, possible since $B\ge k$, has probability $\prod_{i\notin Q}c_i>0$. A
nondecreasing polynomial that takes two different values is strictly increasing, since equal values at two points
would make it constant on an interval, hence everywhere. Every $p\in\mathcal P_{Q,c}$ satisfies
$p\le(\one_Q,c)\in\mathcal P_{Q,c}$, which gives the supremum. (b) follows from (a). (c) At a finite $u$ with
$F^Q_B(u)=\epsilon<\bar\epsilon_Q$, the polynomial $\lambda\mapsto V_B(p(u)+\lambda(\one_Q-p(u)),c)$, which moves the
queried coordinates only, is nondecreasing, equals $\epsilon$ at $0$ and $\bar\epsilon_Q$ at $1$, and is therefore
strictly increasing. Since $u+\delta\one_Q$ corresponds to $\lambda=1-e^{-\delta}$, every $\delta>0$ gives a point of
$U^Q$, so $S^Q\subseteq\overline{U^Q}$; continuity gives the reverse inclusion. By (a), $x\mapsto F^Q_B(x\one_Q)$ is
continuous and strictly increasing from $0$ to $\bar\epsilon_Q$, so $t_Q$ exists and is unique. Attainment and
$\I^Q_*\le t_Q$ follow as in Lemma~\ref{lem:closure}.
\end{proof}

\begin{lemma}[Restricted information bounds]\label{lem:rzeropath}
Let $B\ge k$ and $0<\epsilon<\bar\epsilon_Q$. Every evaluator valid over $\mathcal P_{Q,c}$ satisfies
$\E_0N\ge c_\alpha/\I^Q_*$; if $T=\E_0N<\infty$ and $w_i=\E_0N_i/T$ for $i\in Q$, then $T\,I^Q_w\ge c_\alpha$. If
$w^*$ attains $\I^Q_*$, the fixed-count all-zero design $n_i=\lceil w^*_ic_\alpha/\I^Q_*\rceil$ ($i\in Q$) is valid
over $\mathcal P_{Q,c}$, so $c_\alpha/\I^Q_*\le\inf\E_0N<c_\alpha/\I^Q_*+|Q|$, and
$t_Q/|Q|\le I^Q_{\one/|Q|}\le\I^Q_*\le t_Q$.
\end{lemma}

\begin{proof}
The proofs of Lemma~\ref{lem:zeropath} and Proposition~\ref{prop:sandwich} use only that every finite $u\in U^Q$ is the
queried part of an unsafe parameter of the class, that the responses depend on the queried coordinates alone, and
coordinatewise monotonicity (Lemma~\ref{lem:envelope}(a)). All three hold here, with Lemma~\ref{lem:rclosure} in place of
Lemma~\ref{lem:closure}. An unsafe parameter with some queried $p_i=1$ either has a positive count at $i$, so that the
all-zero probability is zero, or can be replaced by a large finite value of $u_i$ without leaving $U^Q$, by continuity.
\end{proof}

\subsection{The general cycle-aware certificate}

Before any response, choose a nonempty set $A$ of segments and a set
$Q_A\subseteq Q\cap E_A$ of milestones to query that meets every segment of $A$; put $K'=|Q_A|$, and bound milestones
outside $Q_A$ by $\tilde c_i$, with $\tilde c_i=c_i$ for $i\notin Q$ and $\tilde c_i=1$ for $i\in Q$. For $j\in A$, let
$\Hc_j$ consist of the $h_j$ milestones of segment $j$ before its first milestone in $Q_A$. Define the mean
truncated cycle length
\[
  \sigma_j=\sum_{h=0}^{h_j}\ \prod_{\substack{i\in\Hc_j\\\text{first }h}}\tilde c_i.
\]
This is the expected number of attempts in one cycle of segment $j$ up to either a prefix failure or the first
attempt at a queried milestone, including the terminating attempt, at the declared constants. Thus $\sigma_j\in[1,h_j+1]$, and $\sigma_j=1$ when the segment
begins with a queried milestone. Next put
\[
  \rho_j=\frac{\prod_{i\in\Hc_j}\tilde c_i}{\sigma_j}.
\]
The numerator is the probability of reaching the first queried milestone; $\rho_j$ is its ratio to the mean
truncated cycle length. Finally, let $\bar c_j=\prod\tilde c_i$, with the product over milestones of segment $j$
outside $Q_A$ that follow its first milestone in $Q_A$. This accounts for the unqueried steps after that milestone.
Define $D_{A,Q_A}(B)$ as the smaller of
\[
  C_A(B)\!\!\prod_{i\in E_A\setminus Q_A}\!\!\tilde c_i
  \quad\text{and}\quad
  \frac{(B-k+K_A)^a}{a!}\prod_{j\in A}\rho_j\bar c_j .
\]

\begin{theorem}[Cycle-aware selected-segment certificate]\label{thm:cyclecert-general}
Let $B\ge k$.
\begin{enumerate}
\item[(i)] For every $p\in[0,1]^k$ with $p\le\tilde c$ coordinatewise, in particular every parameter of the class,
and every process in $\Dcl(p)$, $\Prob(\text{completion within }B)\le V_B(p)\le D_{A,Q_A}(B)\prod_{i\in Q_A}p_i$.
\item[(ii)] If $\epsilon\ge D_{A,Q_A}(B)$, certifying before any response is valid. Otherwise let
$r'=(\epsilon/D_{A,Q_A}(B))^{1/K'}$ and $\tau'=-\log(1-r')$. Querying each $i\in Q_A$ exactly
$n'=\lceil c_\alpha/(K'\tau')\rceil$ times and certifying if and only if all responses are zero is valid over every
class with these $(Q,c)$, certifies with probability one at $p_Q=0$, and uses $N=K'n'<c_\alpha/\tau'+K'$ responses.
If in addition $\epsilon<V_B(\one_Q,c)$, then $\tau'\le\I^Q_*\le t_Q$.
\item[(iii)] If $Q_A=E_A$, the rule is that of Theorem~\ref{thm:finite}. In general, the second term of
$D_{A,Q_A}(B)$ divided by the first equals $\bigl[(B-k+K_A)^a/(a!\,C_A(B))\bigr]\prod_{j\in A}\sigma_j^{-1}$, which
tends to $\prod_{j\in A}\sigma_j^{-1}$ as $B\to\infty$.
\end{enumerate}
\end{theorem}

Part (iii) locates the gain: as $B\to\infty$ the ratio of the cycle-aware coefficient to the window coefficient
tends to $\prod_{j\in A}\sigma_j^{-1}\le1$, and $\sigma_j>1$ only when segment $j$ has an unqueried prefix before its
first queried milestone. Declared constants after that milestone enter both terms in the same way, and when every selected
segment begins with a queried milestone the cycle term is never the smaller one. The proof bounds $\beta_j/\kappa_j$
through Lemma~\ref{lem:renewal}(a) and uses that $x_1\cdots x_h/(1+x_1+x_1x_2+\dots+x_1\cdots x_h)$ is nondecreasing
in each coordinate (proof below).

\begin{proof}[Proof of Theorem~\ref{thm:cyclecert-general}]
(i) The first inequality is Lemma~\ref{lem:envelope}(c). The window term is Theorem~\ref{thm:finite}(i) with
$p_i\le\tilde c_i$ on $E_A\setminus Q_A$. For the cycle term, Lemma~\ref{lem:renewal-full}(b) gives
$V_B(p)\le((B-k+K_A)^a/a!)\prod_A\beta_j/\kappa_j$. Fix $j\in A$ and let $x_1,\dots,x_h$ ($h=h_j$) be the
probabilities of $\Hc_j$ in order. Since the first queried milestone is at position $h<\ell_j$,
$\kappa_j\ge\sum_{h'\le h}\pi_{j,h'}$, so $\beta_j/\kappa_j\le g(x)\prod_{i\in E_j\setminus\Hc_j}p_i$ with
\[
  g(x)=\frac{x_1\cdots x_h}{1+x_1+x_1x_2+\dots+x_1\cdots x_h}.
\]
The function $g$ is nondecreasing in each coordinate on $[0,1]^h$: it vanishes when some $x_i=0$, and otherwise
$1/g(x)=\sum_{m=0}^{h}\prod_{i=m+1}^{h}x_i^{-1}$ is a sum of terms that are nonincreasing in each $x_i$. Hence
$g(x)\le g(\tilde c_{\Hc_j})=\rho_j$, and $\prod_{E_j\setminus\Hc_j}p_i\le\bar c_j\prod_{Q_A\cap E_j}p_i$.

(ii) Since $(\one_Q,c)\in\mathcal P^{\rm C}_{Q,c}$, (i) gives $\bar\epsilon_Q\le D_{A,Q_A}(B)$, so if $\epsilon\ge
D_{A,Q_A}(B)$ the class contains no unsafe parameter and certifying immediately is valid. Otherwise, let $p\le\tilde c$
with $V_B(p)>\epsilon$. By (i), $\prod_{Q_A}p_i>r'^{K'}$, so, as in the proof of Theorem~\ref{thm:finite}(iii),
$\sum_{Q_A}u_i>K'\tau'$ (the all-zero probability is zero if some queried $p_i=1$) and
$\Prob_p(\text{all zero})\le e^{-n'K'\tau'}\le\alpha$. Because $(A,Q_A)$ is fixed before the data, this is the
certification probability; at $p_Q=0$ every response is zero, and $\lceil x\rceil<x+1$. If $\epsilon<\bar\epsilon_Q$,
then for $u\in S^Q$, (i) at $(p(u),c)$ gives $\epsilon\le D_{A,Q_A}(B)\prod_{Q_A}(1-e^{-u_i})$, and the concavity of
$u\mapsto\log(1-e^{-u})$ gives $K'^{-1}\sum_{Q_A}u_i\ge\tau'$, so $\tau'\le I^Q_{w'}$ for $w'$ uniform on $Q_A$; the
remaining inequalities are Lemma~\ref{lem:rclosure}(c).

(iii) Direct substitution; for the ratio, $\prod_{E_A\setminus Q_A}\tilde c_i=\prod_A\rho_j\sigma_j\bar c_j$.
\end{proof}

When every selected segment begins with a queried milestone ($h_j=0$ for $j\in A$), the cycle term is at least the
window term at every $B$, because $a!\,C_A(B)=(B-k+a)\cdots(B-k+1)\le(B-k+K_A)^a$.

\begin{proof}[Proof of Theorem~\ref{thm:cyclecert}]
Apply Theorem~\ref{thm:cyclecert-general} with $Q=\{1,\dots,k\}$, so that every unqueried milestone is queryable but
unqueried and $\tilde c\equiv1$. Then $\sigma_j=h_j+1$, $\rho_j=(h_j+1)^{-1}$, $\bar c_j=1$ and
$\prod_{i\in E_A\setminus Q_A}\tilde c_i=1$, so $D_{A,Q_A}(B)=D_B$. Part (i) is Theorem~\ref{thm:cyclecert-general}(i)
for every $p\in[0,1]^k$. Part (ii) is Theorem~\ref{thm:cyclecert-general}(ii): validity over the unrestricted class
implies validity over every subclass, for instance one in which the unqueried steps are known to succeed; the rule
certifies surely whenever the queried probabilities are zero, because then every response is zero; and the
immediate-certification case does not arise, because $\epsilon<1=V_B(\one)\le D_B$ by (i). Part (iii) is
Theorem~\ref{thm:cyclecert-general}(iii) together with the inequality just stated.
\end{proof}

\subsection{Restricted sharp complexity}

Section~\ref{sec:restricted} summarizes the following theorem, with $\Gamma_A$ written as the product of the declared
constants of the active segments times $\prod_{j\in A}\sigma_j^{-1}$, which equals $\prod_{j\in A}\rho_j\bar c_j$ when
$Q_A=Q\cap E_A$.

\begin{theorem}[Restricted sharp complexity and limiting allocation]\label{thm:restricted-full}
Fix the segment lengths, $Q$, $c$ and a nonempty set $A$ of segments, and let $\nu_j=|Q\cap E_j|$. Let integers
$B_n\ge k$ tend to infinity, $p_n\in(0,1)$ tend to zero, $\alpha_n\in(0,1)$, and $t_n=-\log(1-p_n)$, and assume
\begin{enumerate}
\item[(R1)] $\nu_j\ge1$ and $B_np_n^{\nu_j}\to0$ for every $j\in A$;
\item[(R2)] for every $j\notin A$, either $\nu_j=0$ or $B_np_n^{\nu_j}\to\infty$;
\item[(R3)] $c_{\alpha_n}/t_n\to\infty$.
\end{enumerate}
Let $\epsilon_n=V_{B_n}(p_n\one_Q,c)$, $K'_A=\sum_{j\in A}\nu_j$, and $\Gamma_A=\prod_{j\in A}\rho_j\bar c_j$ with
$Q_A=Q\cap E_A$. Then $\epsilon_n\to0$, $t_n$ is the restricted diagonal threshold at $(B_n,\epsilon_n)$,
$\epsilon_n\sim(B_n^a/a!)\,\Gamma_A\,p_n^{K'_A}$, and for every class with these $(Q,c)$, whether fixed, capped or
mixed,
\[
  \inf_{\text{valid over the class}}\E_0N\;\sim\;c_{\alpha_n}\Bigl(\frac{B_n^a\,\Gamma_A}{a!\,\epsilon_n}\Bigr)^{1/K'_A},
\]
attained by the certificate of Theorem~\ref{thm:cyclecert-general} with $(A,Q\cap E_A)$. For every sequence of evaluators
valid over the class with $\E_0N\sim c_{\alpha_n}/t_n$, the allocation weights $\E_0N_i/\E_0N$ tend to $1/K'_A$ on
$Q\cap E_A$ and to $0$ on $Q\setminus E_A$.
\end{theorem}

\begin{proof}[Proof of Theorem~\ref{thm:restricted-full}]
\emph{Risk asymptotics and threshold.} On the restricted diagonal, for $j\in A$,
$\beta_j=p_n^{\nu_j}\prod_{E_j\setminus Q}c_i$, so $B_n\beta_j\to0$ by (R1), and $\kappa_j\to\sigma_j$ because
$\pi_{j,h}$ contains a factor $p_n$ exactly when $h>h_j$. Hence $\beta_j/\kappa_j\sim\rho_j\bar c_jp_n^{\nu_j}$. For
$j\notin A$, $\kappa_j\le\ell_j$, and $\beta_j$ is a positive constant if $\nu_j=0$ and of exact order $p_n^{\nu_j}$
otherwise, so (R2) gives $\kappa_j/(B_n\beta_j)\to0$. Lemma~\ref{lem:renewal-full}(d) gives the asymptotics of
$\epsilon_n$, which tends to zero because $B_n^ap_n^{K'_A}=\prod_{j\in A}B_np_n^{\nu_j}$. By
Lemma~\ref{lem:rclosure}(a), $\epsilon_n=F^Q_{B_n}(t_n\one_Q)<\bar\epsilon_Q$, and by uniqueness $t_Q=t_n$.

\emph{Information.} Write $D_n=D_{A,Q\cap E_A}(B_n)$. Its cycle term is $\sim(B_n^a/a!)\Gamma_A$, and its window term
is $C_A(B_n)\prod_A\rho_j\sigma_j\bar c_j\ge(1+o(1))(B_n^a/a!)\Gamma_A$ since $\sigma_j\ge1$; hence
$D_n\sim(B_n^a/a!)\Gamma_A$. Theorem~\ref{thm:cyclecert-general}(i) gives $\epsilon_n\le D_np_n^{K'_A}<D_n$, and by the risk
asymptotics $r'_n=(\epsilon_n/D_n)^{1/K'_A}\sim p_n$, so $\tau'_n\sim t_n$, while Theorem~\ref{thm:cyclecert-general}(ii) gives
$\tau'_n\le\I^Q_{*,n}\le t_n$. Thus $\tau'_n\sim\I^Q_{*,n}\sim t_n\sim(a!\,\epsilon_n/(B_n^a\Gamma_A))^{1/K'_A}$.

\emph{Complexity.} By Lemma~\ref{lem:rzeropath} and Theorem~\ref{thm:cyclecert-general}(ii),
$c_{\alpha_n}/\I^Q_{*,n}\le\inf\E_0N<c_{\alpha_n}/\tau'_n+K'_A$, and the additive $K'_A$ is negligible by (R3). The
validity constraint, hence the infimum, is the same for every class with these $(Q,c)$.

\emph{Allocation.} Fix $z\in(0,\infty)^Q$. The parameters with $1-e^{-t_nz_i}\sim p_nz_i$ on $Q$ and $c$ off $Q$
satisfy the hypotheses of Lemma~\ref{lem:renewal-full}(d) with the same $A$ (the products $B_n\beta_j$ change by bounded
factors), and their active means $\kappa_j$ have the same limits $\sigma_j$, because prefixes contain no queried
coordinate. Hence $F^Q_{B_n}(t_nz)/\epsilon_n\to\prod_{i\in Q\cap E_A}z_i$. The proof of Theorem~\ref{thm:alloc} then
applies verbatim with $E_A$, $K_A$, $\Delta_k$, $F_{B_n}$ and Lemma~\ref{lem:zeropath} replaced by $Q\cap E_A$, $K'_A$,
$\Delta_Q$, $F^Q_{B_n}$ and Lemma~\ref{lem:rzeropath}; the coordinates in $Q\setminus E_A$ play the role of the free
coordinates.
\end{proof}

\begin{example}[A queried milestone in a free segment]\label{ex:freeq}
Take segments of lengths $(2,3)$ with $p=(1,q\mid1,q,q)$, $Q=\{2,4,5\}$ and $p_1=p_3=1$ fixed, along $q=p_n=B_n^{-3/4}$.
Segment~1 has $\nu_1=1$ and $B_nq\to\infty$, so it is free although it contains a queried milestone. Segment~2 has
$\nu_2=2$ and $B_nq^2\to0$, so $A=\{2\}$ with $h=1$, $\sigma=2$, $\rho=1/2$, $\bar c=1$, $\Gamma_A=1/2$ and
$K'_A=|Q\cap E_A|=2$, not $|Q|=3$. Theorem~\ref{thm:restricted-full} gives $\epsilon_n\sim B_nq^2/2$ and
$\inf\E_0N\sim c_\alpha(B_n/(2\epsilon_n))^{1/2}$, attained by querying milestones $4$ and $5$ only, with limiting
allocation $(0,1/2,1/2)$ on $Q$. Convergence is slow because the free segment enters through
$\kappa_1/(B_n\beta_1)=2B_n^{-1/4}$: numerically, $\epsilon_n/(B_nq^2/2)=0.658,\,0.799,\,0.887,\,0.937$ and
$\tau'_n/t_n=0.812,\,0.894,\,0.942,\,0.968$ at $B_n=10^3,10^4,10^5,10^6$.
\end{example}

\begin{proof}[Proof of Corollary~\ref{cor:six}]
The window coefficient $B-5$ is Theorem~\ref{thm:finite}(i) with the single segment and the sure transitions capped at
one. The cycle coefficient is Theorem~\ref{thm:cyclecert}(i) with $h=1$, $K_A=6$ and $a=1$ (equivalently,
Theorem~\ref{thm:cyclecert-general} with $Q_A=\{2,3,4\}$, $\sigma=2$, $\rho=1/2$ and $\bar c=1$), so the cycle term is
$B/2$, which is at most $B-5$ exactly when $B\ge10$. The fixed and capped versions of the class pose the same problem
(Appendix~\ref{app:classes}). The asymptotics are Theorem~\ref{thm:restricted-full} with $A$ the single segment,
$\nu_1=3$, $\Gamma_A=1/2$ and $K'_A=3$; for the window certificate, $r/t_n\sim(\epsilon_n/B_n)^{1/3}/p_n\to2^{-1/3}$.
\end{proof}

\begin{remark}[Scope of the restricted results]\label{rem:rscope}
(a) Cycle lengths change the restricted constant only through the factors $\sigma_j^{-1}$, that is, only when an active
segment has fixed, capped or unqueried milestones before its first queried one; declared constants after the first
queried milestone enter through $\bar c_j$, exactly as in the window bound. (b) A segment without queryable milestones
is always free. (c) At a finite budget the optimal weights need not be uniform (compare Example~\ref{ex:nonuniform});
only limiting ratios of expected counts are characterized. The asymptotic optimality result does not cover crossovers
($B_np_n^{\nu_j}\to\gamma\in(0,\infty)$ for some $j$), fixed $\epsilon$, declared constants that vary with $n$,
efficiency at nonzero safe parameters, or realized adaptive allocations. These exclusions do not affect the validity
of the finite certificates.
\end{remark}

\section{Numerical methods and reproduction}\label{app:numerics}

\paragraph{Exact recursion.} $V_B(p)$ is computed by the forward recursion over the states $0,\dots,k$ for $B$ steps,
in exact rational arithmetic or in binary64. For large budgets the same quantity is computed as the $(0,k)$ entry of
the one-step transition matrix raised to the power $B$ by repeated squaring in binary64; all entries are nonnegative,
so no cancellation occurs. The two agree to a relative difference of $1.3\times10^{-15}$ against exact rationals on
four test cases and of $5\times10^{-14}$ against the binary64 recursion on four cases with $B\le2000$.

\paragraph{Thresholds.} Diagonal and restricted diagonal thresholds are found by bisection on
$x\mapsto V_B(x\one)$, or on $x\mapsto V_B(1,x,x,x,1,1)$ for the six-transition family, with 64 halvings. The thresholds in
Table~\ref{tab:six} lie inside exact rational root brackets of width $10^{-12}$ in which the sign change of
$V_B-\epsilon$ was verified in rational arithmetic.

\begin{figure}[ht]
  \centering
  \includegraphics[width=0.6\textwidth]{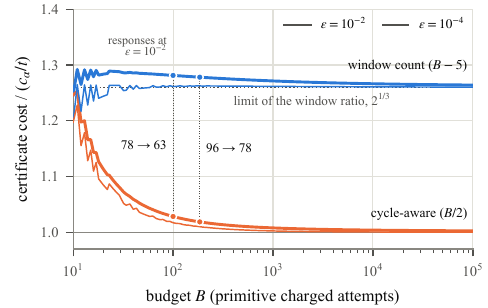}
  \caption{Unrounded certificate cost relative to the lower benchmark, $t/\tau$ (window coefficient $B-5$) and
  $t/\tau'$ (cycle-aware coefficient $B/2$), for the six-transition family at $\alpha=1/40$ and $\epsilon=10^{-2}$
  (thick) or $10^{-4}$ (thin); dots mark the budgets of Table~\ref{tab:six}. The limits $2^{1/3}$ and $1$ hold along
  phase-interior sequences; at fixed $\epsilon$ the curves approach values near them. The oscillation at small $B$ is a
  parity effect: most failed cycles have length two.}
  \label{fig:cycle}
\end{figure}

\paragraph{Figures.} Figure~\ref{fig:complexity} uses 118 distinct integer budgets, log-spaced from $4$ to $10^6$,
together with $B=285$, at $\epsilon=10^{-4}$ and $\alpha=1/40$. At each budget the displayed certificate is the
smallest of the certificates of Theorem~\ref{thm:finite} for all nonempty sets of segments and of
Proposition~\ref{prop:crosscert} for all pairs $(A,M)$; all are computed from $(B,\epsilon,\alpha)$ before any
response, so the minimum is a valid prospective design. Panel (a) plots, at the geometric mean of consecutive budgets
$B_i<B_{i+1}$, the exponent $\log(t_i/t_{i+1})/\log(B_{i+1}/B_i)$ of the lower benchmark; on the final interval ending at $B=10^6$ it is $0.2506$,
$0.3340$, $0.5001$ and $1.0000$ for full reset, segments $(3,1)$, segments $(2,2)$ and full retention.

The profile $h_a$ and its elasticity are evaluated by their
power series for $\lambda\le2$ and by the finite expression
$h_a(\lambda)=\sum_{i=0}^a\binom ai(-1)^ii!\,\lambda^{-i}\Prob(\Gamma_{i+1}\le\lambda)$ otherwise, with $\Gamma_n$ a
standard gamma variable; the two agree on $[1,3]$ to $5\times10^{-15}$ for $h_a$ and $4\times10^{-14}$ for $\lambda h_a'$, and $\tau_\times$ is found by bisection.

On the plotted budget grid, for segments $(3,1)$, Theorem~\ref{thm:finite} alone is up to $18.1\%$ above $c_\alpha/t$ (at $B=279$), and the minimum
with Proposition~\ref{prop:crosscert} is at most $2.8\%$ above it for $B\ge30$ and $6.9\%$ at smaller budgets, where
rounding dominates. Proposition~\ref{prop:crosscert} also lowers the full-retention certificate by up to $1.1\%$ and
that of segments $(2,2)$ by up to $0.3\%$; it does not apply to full reset, which has one segment. Among the certificates of Theorem~\ref{thm:finite} for
segments $(3,1)$, querying only the length-three segment has the smaller unrounded cost $c_\alpha/\tau_A$ from $B=285$
on (the dot in Figure~\ref{fig:complexity}(b)); with rounded counts the minimizing set alternates between $B=277$ and $B=285$ and is
the length-three segment alone from $B=286$ on.

Figure~\ref{fig:cycle} uses 121 log-spaced integer budgets from $10$ to $10^5$,
which include $B=100$ and $B=185$, at $\epsilon\in\{10^{-2},10^{-4}\}$; its rounded counts at $B=100$ and $185$ are
those of Table~\ref{tab:six}.

\paragraph{Table~\ref{tab:gdm}.} The bracket $[0.0880973028788023,\,0.0880973028788024]$ contains
$1-40^{-1/40}$, which is verified by comparing $(1-x)^{40}$ with $1/40$ in exact rationals at both endpoints. At these
endpoints exact rational evaluation gives, under full retention, $V_{10}<1/100$ at the upper endpoint and
$V_{11}>1/100$ at the lower one; under full reset, $V_{186}>1/100$ at the lower endpoint $x$, where $-\log(1-x)\one\in\Pi$; and in
whole-trial units, $1-(1-\prod_ip_i)^{166}<1/100$ at the upper endpoint and $1-(1-\prod_ip_i)^{167}>1/100$ at the
lower one. The window bounds quoted in Section~\ref{sec:gdm} are also exact: under full reset $166\,x^4<1/100$ at the
upper endpoint and $167\,x^4>1/100$ at the lower one, which gives the analytic budget $B=169$; under full retention
$\binom94x^4<1/100<\binom{10}4x^4$ at the appropriate endpoints, which gives 9.

The maximizer of $F_{185}$ over $\Pi$ is not the symmetric point. Consider the rational point
\[
  q=\Bigl(\frac{831115217}{10^{10}},\ \frac{223426761}{2500000000},\
  \frac{449584573}{5000000000},\ \frac{35988607}{400000000}\Bigr).
\]
It satisfies $\prod_i(1-q_i)^{10}\ge1/40$, so $-\log(\one-q)\in\Pi$. Exact evaluation gives $V_{185}(q)>V_{185}(x\one)$ at the upper
endpoint $x$ ($0.0099791$ against $0.0099549$), hence $V_{185}(q)$ exceeds the value at the symmetric point. The point was
located by a binary64 search; only the exact comparison is used.

The upper bound on full-reset risk at $B=185$ is established by a
branch and bound over the region $\{u\ge0:\sum_iu_i\le\log(40)/10\}$ in integer arithmetic at scale $2^{31}$: a rational upper bound on the domain cap $\log(40)/10$ is
certified by an exact series comparison, $e^{-x}$ is enclosed by alternating partial sums with directed integer
rounding, every product in the recursion is rounded up, coordinatewise monotonicity bounds a box by its upper corner,
and a box is accepted when its bound is at most $\lfloor2^{31}/100\rfloor-\lceil2^{31}/10^9\rceil=21474833$. The run
evaluates the recursion $2{,}568{,}639$ times, accepts $1{,}284{,}320$ boxes at depth at most $38$, and gives
$\sup_{u\in\Pi}F_{185}(u)\le21474833/2^{31}<1/100-10^{-9}$. The cutoff was fixed before the search: its gap of about
$1.6\times10^{-9}$ below $1/100$ is not the unknown distance between the worst-case risk and $1/100$.

A separate implementation at scale $2^{60}$, in C++, gives the same
conclusion with $2{,}557{,}015$ evaluations and writes its $1{,}278{,}508$ accepted boxes as a list of dyadic paths
with their bounds.

The supplement contains an independent checker in standard-library Python, written
from the format and the bound's specification, which treats that list as untrusted. It proves the domain cap in exact
rationals, checks that the paths tile the root box, rebuilds every box, and recomputes the bound of every leaf at its
upper corner; all $1{,}278{,}508$ recomputed bounds are at most the acceptance cutoff and equal the recorded ones, the
largest being $11529213865092556/2^{60}$. A successful check certifies $\sup_{u\in\Pi}F_{185}(u)<1/100-10^{-9}$ whichever
program produced the list, so the trusted computational base of this certificate is the checker, the integer
arithmetic and decompression of the Python interpreter, and the monotonicity and rounding arguments stated in the
checker (Lemma~\ref{lem:domination}, alternating Taylor enclosures of $e^{-x}$ with outward rounding, and products
rounded up in a recursion with nonnegative coefficients); neither the C++ compiler nor a numerical library is needed.
No floating-point number influences an acceptance decision, and the certificate is not verified in a proof assistant.

\paragraph{Sensitivity of the reset-versus-retention comparison.} Table~\ref{tab:sensitivity} changes one parameter at
a time from the setting of Section~\ref{sec:gdm}: $k$ milestones, $n$ zero responses at each, risk level $\epsilon$ and
confidence level $1-\alpha$. The plausibility region of Proposition~\ref{prop:plausibility} has symmetric point
$\bar p=1-\alpha^{1/(nk)}$. The three budgets in each row are of different kinds. The full-reset certificate is
analytic: the largest $B$ with $(B-k+1)\bar p^k\le\epsilon$ (Theorem~\ref{thm:finite}), checked in exact rationals at
the end points of a rational bracket of $\bar p$. The full-reset limit is an exact risk calculation, not a
certificate: at the next budget $V_B(\bar p\one)>\epsilon$, and $-\log(1-\bar p)\one\in\Pi$, so the plausibility rule cannot certify that budget. The largest budget certified
by this rule lies between the analytic certificate and this limit (185 in the first row). The
full-retention budget is an exact certificate: by Proposition~\ref{prop:schur} the supremum over $\Pi$ is attained at
$-\log(1-\bar p)\one$, where the risk is enclosed by the recursion in integer fixed point (scale $2^{256}$) with every product
rounded outward. No branch and bound was run for this table.

Full reset certifies more attempts than full retention in
every row, by a factor between $2.0$ ($k=3$) and about $7{,}200$ ($k=6$). The factor grows with $k$, $n$ and
$\epsilon$: the full-reset certificate is about $\epsilon/\bar p^k$, whereas the full-retention budget is of order
$(k!\,\epsilon)^{1/k}/\bar p$. In the displayed cases with fewer milestones, fewer responses or a stricter risk level,
the factors are approximately 2--3.2.

\begin{table}[ht]
  \centering
  \caption{Largest certified budget $B$ (primitive charged attempts) from $n$ zero responses at each of $k$ milestones,
  one parameter changed at a time; Proposition~\ref{prop:plausibility} at level $\alpha$, risk level $\epsilon$. Full
  reset: analytic window certificate, and the limit set by the exact risk at the symmetric point (no budget certified by this rule can
  exceed it). Full retention: exact certificate. Ratio: full-reset certificate over full-retention certificate.
  Reproduced by \texttt{sensitivity.py}.}
  \label{tab:sensitivity}
  \small
  \begin{tabular}{lrrrrrrrrr}
    \toprule
     & & & & & & \multicolumn{2}{c}{full reset} & full retention & \\
    \cmidrule(lr){7-8}\cmidrule(lr){9-9}
    setting & $k$ & $n$ & $\epsilon$ & $\alpha$ & $\bar p$ & certificate & limit & certificate & ratio\\
    \midrule
    baseline & 4 & 10 & $1/100$ & $1/40$ & 0.0881 & 169 & 185 & 10 & 16.9\\
    $n=5$ & 4 & 5 & $1/100$ & $1/40$ & 0.1684 & 15 & 17 & 6 & 2.5\\
    $n=20$ & 4 & 20 & $1/100$ & $1/40$ & 0.0451 & 2{,}427 & 2{,}554 & 19 & 127.7\\
    $n=40$ & 4 & 40 & $1/100$ & $1/40$ & 0.0228 & 37{,}061 & 38{,}116 & 37 & 1{,}001.6\\
    $k=3$ & 3 & 10 & $1/100$ & $1/40$ & 0.1157 & 8 & 9 & 4 & 2.0\\
    $k=6$ & 6 & 10 & $1/100$ & $1/40$ & 0.0596 & 222{,}455 & 237{,}751 & 31 & 7{,}176.0\\
    $\epsilon=1/1000$ & 4 & 10 & $1/1000$ & $1/40$ & 0.0881 & 19 & 21 & 6 & 3.2\\
    $\epsilon=1/10$ & 4 & 10 & $1/10$ & $1/40$ & 0.0881 & 1{,}663 & 1{,}920 & 20 & 83.2\\
    $\alpha=1/20$ & 4 & 10 & $1/100$ & $1/20$ & 0.0722 & 371 & 402 & 12 & 30.9\\
    \bottomrule
  \end{tabular}
\end{table}

\paragraph{Diagnostics.} The numbers in Examples~\ref{ex:nonuniform} and~\ref{ex:freeq} are binary64 evaluations
(golden-section search over weights with bisection for the boundary, and the exact recursion along the stated
sequences); they illustrate the statements and prove nothing. For the six-transition family at $B=100$ and
$\epsilon=1/100$, the restricted information at the normalized-gradient weights $(0.327,0.336,0.337)$, computed by a
Nelder--Mead search with bisection, equals $t$ to six digits, which suggests $\I^Q_*\approx t$ there; the fixed-count
design of Lemma~\ref{lem:rzeropath} with these weights would use $62$ responses. This is a numerical observation and
not a certificate.

\paragraph{Supplement.} The self-contained supplement, available from the authors on request, reproduces the
numerical results, figure data and exact-arithmetic checks described above. Its full checker independently
recomputes all $1{,}278{,}508$ leaf bounds of the $B=185$ certificate; commands, dependencies and runtime guidance
are in its README.

\end{document}